%% file: main.tex
\documentclass[letterpaper]{article} 
\usepackage[preprint]{aaai2027}
\usepackage[hyphens]{url}  
\usepackage{graphicx} 
\usepackage{natbib}  
\usepackage{caption} 
\usepackage{algorithm}
\usepackage{algorithmic}

\usepackage{enumitem}
\usepackage{siunitx}

\usepackage{booktabs}
\usepackage{pifont}     
\usepackage{colortbl}   
\usepackage[table]{xcolor}
\usepackage{amsmath} 
\usepackage{multirow}
\usepackage{amssymb}
\usepackage{xcolor}
\usepackage[most]{tcolorbox}
\usepackage{rotating}

\usepackage{newfloat}
\usepackage{listings}
\DeclareCaptionStyle{ruled}{labelfont=normalfont,labelsep=colon,strut=off} 
\floatstyle{ruled}
\newfloat{listing}{tb}{lst}{}
\floatname{listing}{Listing}

\usepackage{booktabs}

\title{Recalling the Past to Answer the Present: Context-Aware Multi-Evidence Reasoning for Streaming Video Question Answering}
\author{
    Cui Yakun\textsuperscript{1},
    Tiantian Geng\textsuperscript{2},
    Huixiang Kong\textsuperscript{3},\\
    Xingqun Qi\textsuperscript{4},
    Sirui Han\textsuperscript{1},
    Yike Guo\textsuperscript{1}
}
\affiliations{
    \textsuperscript{1}The Hong Kong University of Science and Technology\\
    \textsuperscript{2}University of Birmingham\\
    \textsuperscript{3}University of Cambridge\\
    \textsuperscript{4}North China Electric Power University\\
    ycuibd@connect.ust.hk
}

\begin{document}

\maketitle
\begin{figure*}[t]  
    \centering
    \includegraphics[width=\textwidth]{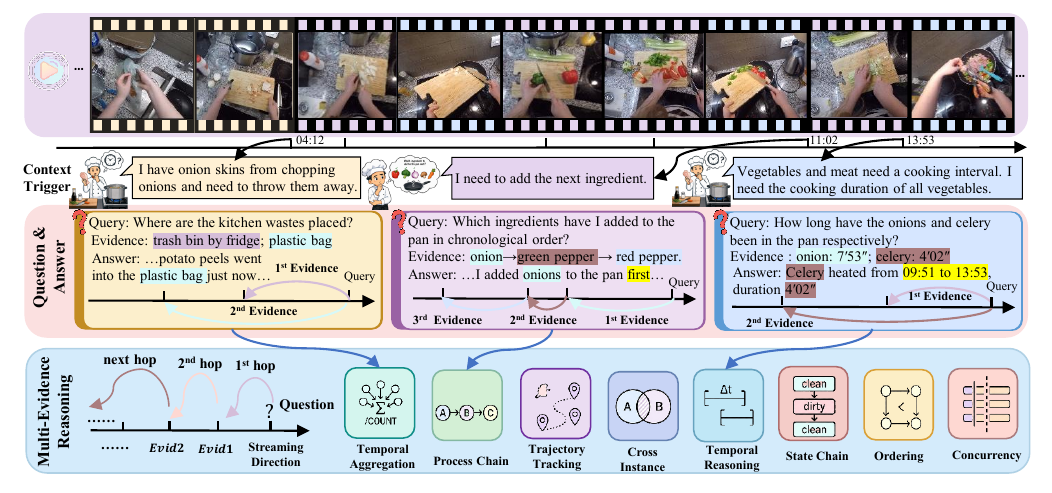}  
    \captionsetup{justification=centering, singlelinecheck=false}
    \caption{Details for multi evidence reasoning in streaming video \& context-aware VQA \& 8 multi evidence reasoning patterns.} 
    \label{bench_detail} 
\end{figure*}

\begin{abstract}
Vision-language models (VLMs) have achieved remarkable progress in video understanding, yet existing video question answering (VQA) benchmarks mainly evaluate models in offline settings. Real-world streaming scenarios instead require models to answer context-dependent queries by retrieving and reasoning over evidence distributed across the video history.
We introduce StreamRecall, a benchmark for context-aware Multi evidence reasoning in streaming video QA. StreamRecall requires models to retrieve, associate, and reason over temporally scattered evidence to answer queries at the appropriate moment. It contains 4,983 questions over 462 videos, with context-aligned query times and fine-grained temporal evidence annotations. We further propose the Evidence Utilization Cascade to diagnose model reasoning through evidence localization, grounding, chain completion, and answer derivation.
To enable efficient long-term streaming reasoning, we present EviStream, a training-free causal visual memory mechanism that preserves reasoning-critical evidence under a fixed KV budget. Experiments reveal substantial gaps between current VLMs and realistic streaming requirements, establishing StreamRecall as a foundation for developing evidence-aware streaming video intelligence.
\end{abstract}


\section{Introduction}
\label{sec:intro}
Vision-language models (VLMs)~\cite{yang2025qwen3,wang2025internvl3,team2023gemini,singh2025openai,tang2025video} have recently achieved remarkable progress in video understanding~\cite{yang2025vca,zhou2025mlvu}, enabling models to interpret complex visual content through natural language interaction. Consequently, numerous video question answering (VQA) benchmarks have been introduced to evaluate their video comprehension capabilities~\cite{cheng2024videollama,li2024mvbench,wang2025lvbench,fang2024mmbench,wu2024longvideobench}. However, these benchmarks typically assume an offline setting, where the complete video is available before answering questions. Such an assumption differs fundamentally from real-world interactive scenarios, where an intelligent system must continuously perceive incoming video streams and respond based only on causally available observations.

Recent studies have begun to explore streaming video question answering~\cite{lin2026streamingbench,di2025streaming,xiong2025streaming,yao2025timechat}, aiming to evaluate models under online video processing scenarios. However, existing streaming benchmarks mainly focus on questions that can be answered from a locally observable context, where the required evidence appears within a single recent segment. This setting overlooks a fundamental requirement of real-world interaction: \textbf{users often raise questions at moments when specific information is needed, while the relevant evidence may have occurred much earlier in the video stream.} For example, when a user is preparing a meal, they may ask where the knife is located immediately before cutting vegetables. Answering this question requires the model to recall when the knife was previously placed, associate it with the current activity, and provide a timely response. Therefore, realistic streaming video understanding requires not only online perception, but also \textbf{the ability to retrieve and reason over temporally distributed evidence.}
We refer to this capability as \textbf{context-aware Multi evidence reasoning in streaming videos.} Unlike conventional streaming VQA that relies on recent observations, this setting requires models to identify relevant historical events, associate multiple evidence moments, and perform reasoning across long temporal intervals. The complexity increases as evidence events may be separated by minutes or even tens of minutes, and their relevance is determined not only by visual similarity but also by temporal, semantic, and causal relationships. However, existing streaming benchmarks do not evaluate this capability. They largely emphasize single-step reasoning and lack fine-grained temporal evidence annotations, making it difficult to determine whether model failures originate from missing evidence, incorrect evidence utilization, incomplete reasoning chains, or unsupported answer generation.

Beyond evaluation, enabling such reasoning in practical streaming systems introduces additional computational challenges. As the temporal range of required evidence expands, models need to preserve information from an increasingly large number of video frames. Directly retaining all visual tokens leads to linearly growing KV-cache consumption and repeatedly attends to large amounts of irrelevant content, while long-context degradation further limits the effectiveness of transformer-based architectures. Existing token compression strategies are insufficient for streaming Multi evidence reasoning: text-oriented KV eviction methods ignore video-specific redundancy, while visual compression methods based on frame saliency or motion cues may discard static but reasoning-critical information. In Multi evidence reasoning, the importance of a visual token depends not only on its local appearance but also on whether it preserves unique entities, states, or relationships required for future evidence chains.

To address these limitations, we introduce \textbf{StreamRecall}, a new benchmark for context-aware Multi evidence reasoning in streaming video question answering. StreamRecall evaluates whether models can retrieve, associate, and reason over temporally scattered evidence to answer queries at the appropriate moment. To provide a comprehensive analysis of model reasoning behavior, we further propose the Evidence Utilization Cascade, which decomposes Multi evidence reasoning into five progressive stages: \textbf{Locate, Cite, Ground, Complete, and Derive.} This process-oriented evaluation reveals whether a model successfully identifies necessary evidence, uses it in the correct reasoning role, constructs a complete evidence chain, and produces a supported answer.
Furthermore, to address the system-level challenge of efficient long-term streaming reasoning, we propose \textbf{EviStream}, a training-free causal visual memory mechanism that preserves evidence-critical information under a fixed KV budget. EviStream maintains a compact representation of previously observed content by combining semantic boundary detection, anchor memory, robust candidate filtering, Coverage–Novelty Facility Location, and redundancy-aware token merging. With only 7 out of 196 visual tokens retained per frame, EviStream reduces visual KV writes by 85.7\% while achieving state-of-the-art performance on Multi evidence streaming VQA.
Our contributions are summarized as follows:

\begin{itemize}[leftmargin=*,topsep=2pt,itemsep=4pt]

\item \textbf{A context-aware Multi evidence streaming video QA benchmark.}
We introduce StreamRecall, the first benchmark designed to evaluate Multi evidence reasoning in realistic streaming scenarios. StreamRecall contains 4,983 questions over 462 videos (136 hours) collected from EPIC-KITCHENS~\cite{damen2022epickitchens}, Assembly101~\cite{sener2022assembly101}, and Ego4D~\cite{grauman2022ego4d}. The benchmark covers 1–11 evidence hops across 24 scenario-grounded task categories organized into 8 graph-based reasoning patterns. Each query is paired with a context-aligned query time, grounded answer, and complete temporal evidence annotations, resulting in 14,968 evidence items.

\item \textbf{A process-oriented evaluation framework for evidence utilization.}
We propose the Evidence Utilization Cascade, which evaluates five dependent reasoning stages: \textit{Locate} $\rightarrow$ \textit{Cite} $\rightarrow$ \textit{Ground} $\rightarrow$ \textit{Complete} $\rightarrow$ \textit{Derive}. The resulting metrics provide fine-grained diagnosis of whether models fail in evidence retrieval, grounding, reasoning-chain construction, or final answer generation.

\item \textbf{A training-free evidence-preserving streaming memory mechanism.}
We propose EviStream, a causal visual memory framework that maintains representative semantic information and previously uncovered contextual changes under a fixed KV budget. By preserving reasoning-critical evidence rather than only visually salient content, EviStream enables efficient and effective Multi evidence streaming video reasoning.
\end{itemize}

\begin{table*}[t]
\centering
\small
\setlength{\tabcolsep}{3.8pt}
\renewcommand{\arraystretch}{1.15}
\begin{tabular}{@{}l cc c c c r r c c r r r@{}}
\toprule
\textbf{Benchmark} & \textbf{R-Time} & \textbf{R-Content} & \textbf{Hops} & \textbf{Patterns} & \textbf{Full Ev.} & \textbf{Dur.} & \textbf{\#Q} & $\boldsymbol{\rho_q}$ & $\boldsymbol{\rho_e}$ & $\boldsymbol{\tilde{d}_\textbf{last}}$ & $\boldsymbol{\tilde{d}_\textbf{latest}}$ & $\boldsymbol{\tilde{s}_\textbf{ev}}$ \\
\midrule
StreamingBench       & \ding{55} & \ding{55} & 1     & 0 & \ding{55}           & --       & 4,500 & --   & --   & --   & --   & --   \\
E.T.~Bench           & \ding{55} & \ding{55} & 1     & 0 & \ding{55}           & --       & 7,300 & --   & --   & --   & --   & --   \\
S-EMBER              & \ding{55} & \ding{55} & 1     & 0 & \ding{55}           & 388      & 9,448 & --   & --   & --   & --   & --   \\
\midrule
OVO-Bench            & \ding{55} & \ding{55} & 1     & 0 & \ding{55}$^\dagger$ & --       & 1,640 & --   & --   & 5.0  & 0.0  & 20.0 \\
ProactiveVQA         & \ding{55} & \ding{55} & 1     & 0 & \ding{51}           & 48.8     & 3,510 & 1.2  & --   & 8.5$^*$ & 1.0$^*$ & 7.6  \\
RIVER                & \ding{55} & \ding{55} & 1     & 0 & \ding{51}           & 96.5     & 1,810 & 0.31 & --   & 73.1 & 45.8 & 10.0 \\
\midrule
\rowcolor[gray]{0.93}
\textbf{Ours}        & \ding{51} & \ding{51} & \textbf{1--8} & \textbf{8} & \ding{51} & \textbf{136} & \textbf{4,983} & \textbf{3.8} & \textbf{12.0} & \textbf{59.1} & \textbf{4.5} & \textbf{32.5} \\
\bottomrule
\end{tabular}
\caption{Comparison with existing streaming video understanding benchmarks.
\textbf{R-Time}: ecologically valid query timing (query triggered at the moment of real-world need);
\textbf{R-Content}: query content grounded in situated, real-world reasoning demand;
\textbf{Hops}: range of evidence hops per question;
\textbf{Patterns}: number of Multi evidence reasoning patterns;
\textbf{Full Ev.}: complete temporal evidence grounding annotated;
\textbf{Dur.}: total video duration (hours);
\textbf{\#Q}: total questions;
\textbf{$\rho_q$}: question density (questions/min);
\textbf{$\rho_e$}: evidence density (evidence segments/min of evidence span);
\textbf{$\tilde{d}_\mathrm{last}$} / \textbf{$\tilde{d}_\mathrm{latest}$}: median gap from query to earliest/latest evidence (seconds);
\textbf{$\tilde{s}_\mathrm{ev}$}: median evidence span (seconds).
$^\dagger$Only backward-tracing subtasks have temporal grounding.
$^*$No backward gap; query co-occurs with evidence.
}
\label{tab:bench_comparison}
\end{table*}

\section{Related Work}
\label{sec:related_work}

\paragraph{Long-Video Understanding and Evidence-Grounded VideoQA.}
Long-video benchmarks such as EgoSchema, Video-MME, MLVU, and LongVideoBench evaluate temporal retrieval, state tracking, and reasoning across extended recordings~\cite{mangalam2023egoschema,fu2024videomme,zhou2025mlvu,wu2024longvideobench}.
To reduce visual context, methods such as PruneVid and FrameFusion combine token pruning or merging with importance estimation~\cite{huang2025prunevid,fu2025framefusion}.
Evidence-grounded benchmarks further associate answers with supporting video segments~\cite{xiao2024nextgqa}.
However, these settings are predominantly offline: the complete video can be revisited, and evidence selection may be conditioned on the question.
\paragraph{Streaming Video Understanding and Compression.}
Streaming video models process frames causally and retain historical information under bounded resources.
VideoLLM-Online supports temporally aligned interaction, Flash-VStream maintains complementary short- and long-term memories, and StreamingBench evaluates real-time perception and historical retrieval~\cite{chen2024videollmonline,zhang2025flashvstream,lin2024streamingbench}.
Recent methods control memory growth through KV retrieval, segment-level compression, inter-frame redundancy removal, quantization, and hierarchical token pruning~\cite{di2025rekv,chen2026streamkv,yao2025timechatonline,chen2026streamingtom,wang2026stc}.
However, local or fixed-rate compression may discard evidence needed by future queries, which retrieval cannot recover.


\begin{figure*}[t]  
    \centering
    \includegraphics[width=\textwidth]{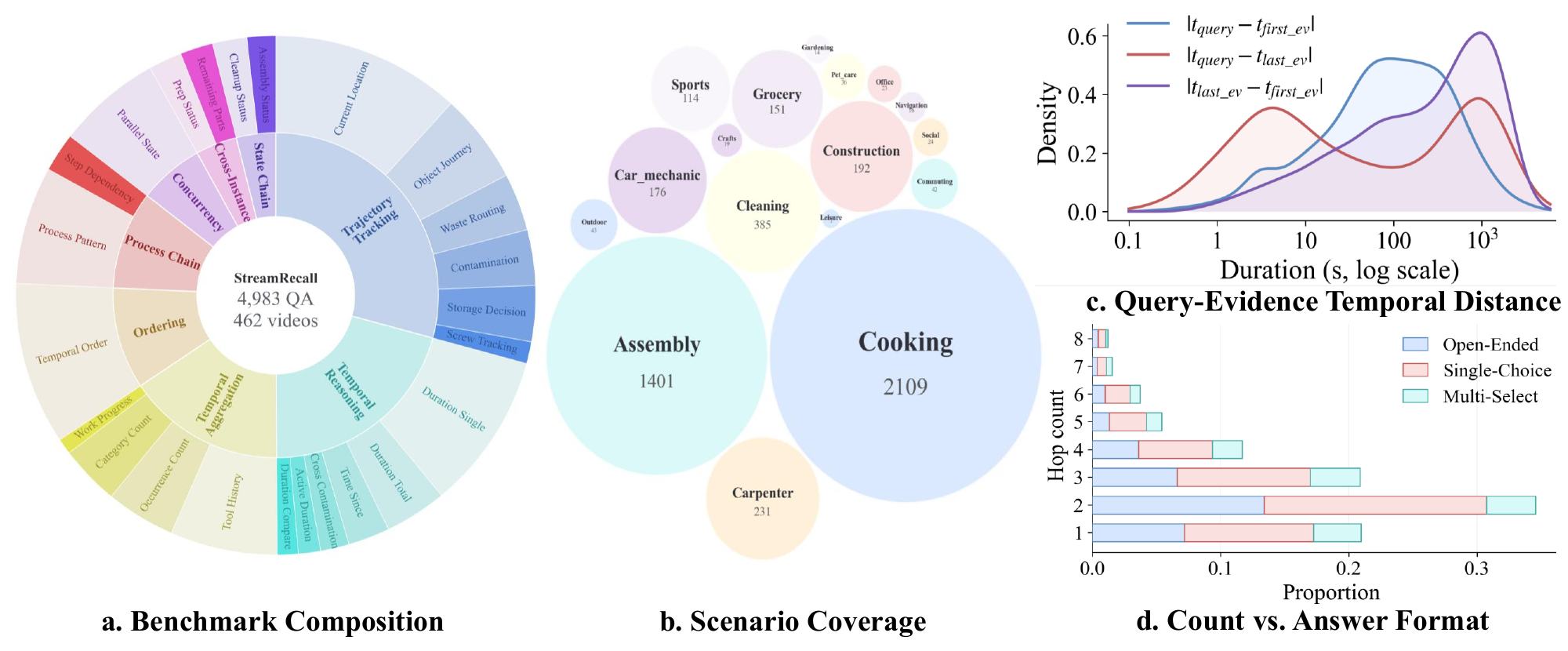}  
    \captionsetup{justification=raggedright, singlelinecheck=false}
    \caption{a. Hierarchical composition of StreamRecall. The inner ring shows dataset source, the middle ring shows reasoning mode, and the outer ring shows QA type. Wedge angles are proportional to sample counts. b. Distribution of QA pairs across 14 activity scenarios. Bubble area encodes sample count. c. Kernel density estimates of temporal distances (log-scaled) across all QA pairs. Three curves show the absolute gap between query time and first evidence, query time and last evidence, and the span from first to last evidence. d. Joint distribution of reasoning depth (evidence count) and answer format. Bar length reflects each evidence value's share of total QA pairs.} 
    \label{bench_detail} 
\end{figure*}

\section{StreamRecall}
\label{sec:benchmark}

Existing streaming video benchmarks are confined to single-hop reasoning without Multi evidence reasoning structure, lack complete temporal evidence grounding, and pose queries that are neither timed nor worded to reflect ecologically valid reasoning demands (Table~\ref{tab:bench_comparison}).
StreamRecall is designed to fill this void: a large-scale, evidence-grounded benchmark that systematically evaluates Multi evidence reasoning under streaming constraints.
We describe its design principles, data construction pipeline, graph-reasoning formalization, and evaluation protocol.

\subsection{Design Principles}
\label{ssec:design}

StreamRecall is built around three interlocking principles that jointly distinguish it from prior work:

\paragraph{Context-Aware Queries.}
Unlike benchmarks that generate synthetic questions at arbitrary timestamps, we impose two constraints to ensure ecological validity during benchmark construction:
\emph{Real Query Time}, which anchors each question to a trigger timestamp $t_\text{query}$ at which a genuine information need arises during viewing;
and \emph{Real Query Content}, which derives each question from a situated reasoning need rather than an arbitrary composition. 


\paragraph{Multi evidence Reasoning.}
Every question requires chaining 1--8 temporally distributed evidence fragments.
We decompose the reasoning structure into 8 graph-reason patterns that capture qualitatively different traversal shapes over a spatial-temporal event graph, enabling fine-grained diagnosis of model capabilities. \textbf{Each QA has one evidence item per reasoning hop: the hop is the reasoning action, and the evidence is its referenced target.}

\paragraph{Streaming-Native Protocol.}
The protocol enforces three causal constraints: (i) frames are processed sequentially, (ii) questions remain unavailable during visual encoding, and (iii) evicted memory cannot be recovered. Evidence may span multiple segments preceding $t_\text{query}$, requiring models to preserve and integrate long-range temporal context.


\begin{figure*}[t]  
    \centering
    \includegraphics[width=\textwidth]{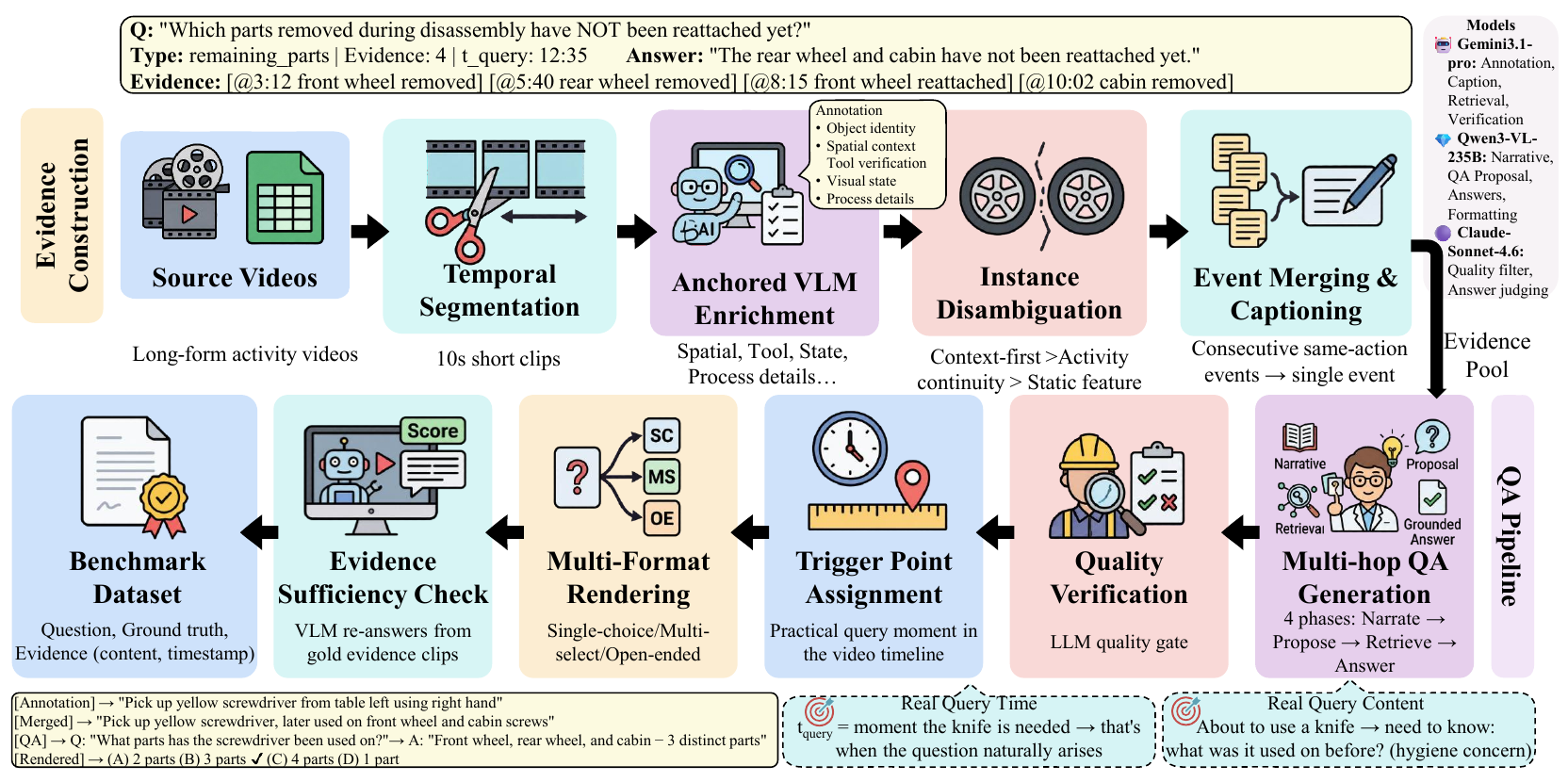}  
    \captionsetup{justification=centering, singlelinecheck=false}
    \caption{Data Construction} 
    \label{bench_data_construc} 
\end{figure*}

\subsection{Data Construction}
\label{ssec:construction}

Figure~\ref{bench_data_construc} summarizes the data construction pipeline.
\textbf{Source Videos and Temporal Segmentation.} We collect 462 videos covering 136 hours and 14 activity scenarios from the validation splits of EPIC-KITCHENS~\cite{damen2022epickitchens}, Assembly101~\cite{sener2022assembly101}, and Ego4D~\cite{grauman2022ego4d}. Each video is divided into clips of at most 10 seconds to support fine-grained VLM analysis. \textbf{Anchored VLM Enrichment.} Guided by the prompt in Figure~\ref{fig:prompt_enrichment}, a VLM generates detailed captions describing object identity, state, spatial context, tool usage, and manipulation details. \textbf{Instance Disambiguation.} Using the prompt in Figure~\ref{fig:prompt_disambiguation}, an LLM resolves repeated object mentions based on activity context and temporal continuity, producing consistent object trajectories. \textbf{Event Merging and Captioning.} Consecutive events involving the same action and object instance are merged to reduce fragmentation, yielding an action chain for each object. \textbf{QA Generation.} Using the prompts in Figure~\ref{fig:prompt_qa_generation}, the pipeline constructs a chronological narrative, proposes questions from 24 QA types, retrieves question-relevant events, and generates evidence-grounded answers. \textbf{Quality Verification.} An LLM verifies each answer against the retrieved event captions and video clips and checks whether the cited events jointly support the answer. The corresponding prompts are shown in Figure~\ref{fig:prompt_verification}. \textbf{Trigger Point Assignment.} A VLM identifies candidate events after all required evidence has appeared, and an LLM selects the point at which the information need most naturally arises. The corresponding prompts are shown in Figure~\ref{fig:prompt_trigger}. \textbf{Manual Annotation.} Human annotators inspect the source videos to verify answer correctness, evidence sufficiency, assigned scene and reasoning-pattern labels.

\subsection{Multi evidence reasoning based on Graph patterns}
\label{ssec:graph}

We formulate Multi evidence streaming video QA as \emph{subgraph reasoning over a spatio-temporal event graph}, providing a unified representation of 8 VQA reasoning patterns.

\paragraph{Event Graph Definition.}
Given a video, we construct a directed graph $\mathcal{G} = (\mathcal{V}, \mathcal{E})$ where each node $v \in \mathcal{V}$ represents an atomic action event characterized by its object, action verb, temporal boundaries $[t_\text{start}, t_\text{end}]$, spatial context, and state information.
Edges $\mathcal{E}$ encode four relation types: \textbf{temporal}, \textbf{same-object}, \textbf{spatial}, and \textbf{causal}.


\paragraph{Reasoning Patterns.}
Each question corresponds to a specific traversal pattern over $\mathcal{G}$.
We identify 8 structurally distinct graph-reasoning modes~\cite{peng2025rewarding, zhang2025lgkgr, shen2025flow}:
\begin{enumerate}[leftmargin=*,topsep=2pt,itemsep=2pt]
\item \textbf{Trajectory Tracking}: following same-object edges to trace an entity's journey across locations.
\item \textbf{Temporal Reasoning}: computing temporal arithmetic (durations, gaps, comparisons) over temporal edges.
\item \textbf{Temporal Aggregation}: selecting nodes by predicate and counting or listing.
\item \textbf{Ordering}: determining the temporal sequence among a set of events via temporal edge traversal.
\item \textbf{Process Chain}: traversing causal and temporal edges to verify action dependencies.
\item \textbf{Concurrency}: identifying co-temporal events through shared time-slice queries.
\item \textbf{Cross-Instance}: reasoning across multiple instances of the same object type via set operations.
\item \textbf{State Chain}: tracking state transitions along same-object edges.
\end{enumerate}
This decomposition supports both diagnosis and method development by enabling per-pattern analysis of task complexity and model performance bottlenecks.

\subsection{Evidence Utilization Cascade}
\label{ssec:evaluation}

Standard answer-accuracy metrics conflate genuine Multi evidence reasoning with fortunate guesses, reasoning shortcuts, and hallucinated evidence.
To diagnose failures in multi-evidence reasoning, we introduce the Evidence Utilization Cascade, a process-oriented evaluation framework.

\paragraph{Cascade Stages.}
The evaluation decomposes reasoning into a dependent pipeline:
\(\text{\textbf{\textit{Locate}}} \rightarrow
\text{\textbf{\textit{Cite}}} \rightarrow
\text{\textbf{\textit{Ground}}} \rightarrow
\text{\textbf{\textit{Complete}}} \rightarrow
\text{\textbf{\textit{Derive}}}\).
Each stage imposes a strictly stronger correctness requirement.
Let $\mathcal{H}$ denote the set of gold reasoning hops for a question, with $|\mathcal{H}| = H$.
For each hop $h \in \mathcal{H}$, we define binary indicators: $\ell_h$ (located), $c_h$ (cited correctly), and $r_h$ (assigned to the correct reasoning role).

\begin{gather}
\text{ER}
=
H^{-1}\sum_{h\in\mathcal{H}}\ell_h,
\qquad
\text{CC}
=
H^{-1}\sum_{h\in\mathcal{H}}\ell_h c_h r_h
\label{eq:hr_cc}
\\[4pt]
\text{CR}
=
(\sum_{h\in\mathcal{H}}\ell_h c_h)
/
(\sum_{h\in\mathcal{H}}\ell_h)
\label{eq:cr}
\\[4pt]
\text{RA}
=
(\sum_{h\in\mathcal{H}}\ell_h c_h r_h)
/
(\sum_{h\in\mathcal{H}}\ell_h c_h)
\label{eq:ra}
\\[4pt]
\text{AF}
=
\mathbf{1}[
\hat{a}\text{ is derivable from }
\{e_h:\ell_h c_h r_h=1\}
]
\label{eq:af}
\end{gather}

\noindent where \textbf{Evidence Recall} (ER) measures evidence localization coverage, \textbf{Citation Recall} (CR) measures factual accuracy among located hops, \textbf{Role Accuracy} (RA) measures whether cited evidence is assigned to the correct reasoning step, \textbf{Chain Completeness} (CC) measures end-to-end per-hop correctness, and \textbf{Answer Faithfulness} (AF) is a binary judgment of whether the final answer follows from the chain.

\paragraph{Auxiliary Metrics.}
We additionally measure \emph{Visual Dependency} (VD)---the degree to which reasoning relies on actual video observations versus prior knowledge ($\text{VD} \in [0,1]$, judged by LLM)---and \emph{Hallucination Rate}.

\paragraph{Cascade Analysis.}
Three meta-metrics characterize error propagation across stages:
\noindent \textbf{Cascade Transition Probability} (CTP) reveals inter-stage bottlenecks; \textbf{First Error Position} (FEP) pinpoints the primary failure stage across the question set $\mathcal{Q}$; \textbf{Error Propagation Rate} (EPR) measures whether upstream errors cascade or are self-correcting.



\begin{figure*}[t]  
    \centering
    \includegraphics[width=\textwidth]{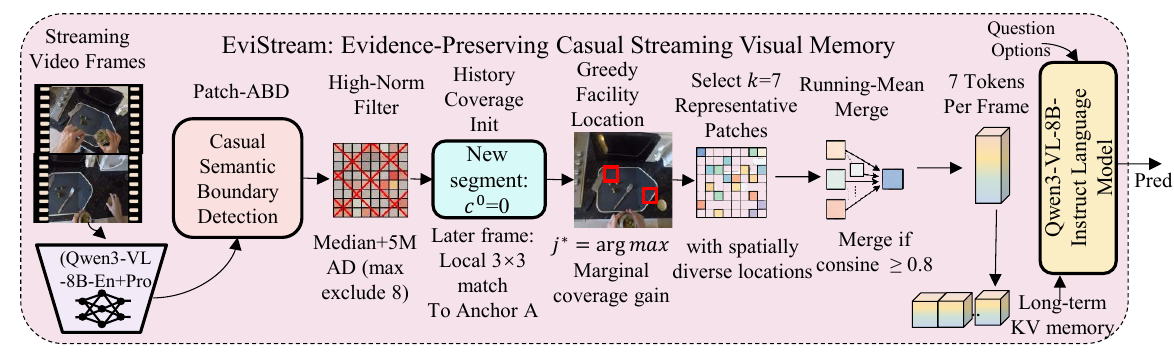}  
    \captionsetup{justification=centering, singlelinecheck=false}
    \caption{structure of EviStream. Further details are provided in Appendix.} 
    \label{model_structure} 
\end{figure*}

\section{EviStream: Evidence-Preserving Causal Visual Memory}
\label{sec:method}

Multi-evidence reasoning over streaming video requires models to assemble causally related evidence fragments scattered across the entire timeline preceding the query trigger. This requirement exposes a limitation of methods that prune KV entries based on visual similarity or temporal redundancy: visually distinct but evidence-free frames may be mistakenly retained as informative.

An effective memory mechanism must be \textbf{causal}~\cite{xu2025beyond,an2026video}, as multi evidence inference relies on logically connected evidence; \textbf{semantic}, as reasoning relevance cannot be reduced to pixel-level change; and \textbf{coverage-preserving}, as any gap in the evidence chain is fatal to Multi evidence inference. We introduce \textbf{EviStream}, a training-free visual memory that causally compresses language-aligned tokens while tracking previously captured scene context.

\subsection{Preliminaries}
\label{ssec:preliminaries}
At time \(t\), a visual encoder \(f_{\mathrm{v}}\) extracts patch representations from the incoming frame \(I_t\), which are mapped by the multimodal projector \(g\) into \(N\) language-aligned visual tokens, \(X_t=\{x_{t,1},\ldots,x_{t,N}\}\), where \(x_{t,i}=g\!\left(f_{\mathrm{v}}(I_t)_i\right)\in\mathbb{R}^{d}\).
These tokens form the visual input to the causal language model.
Each frame yields \(N=14\times14=196\) visual tokens.

\paragraph{Streaming KV Memory.}
During causal inference, the language model stores the keys and values of previously processed tokens in a persistent KV cache \(\mathcal{K}_t\).
Without compression, processing \(T\) frames writes \(NT\) visual tokens, causing the visual KV memory to grow linearly with stream length.
A streaming compressor instead transforms each \(X_t\) into a bounded representation \(\widetilde{X}_t\), with \(|\widetilde{X}_t|\leq k\), and updates the cache as \(\mathcal{K}_t=\operatorname{Update}(\mathcal{K}_{t-1},\widetilde{X}_t)\).
The objective is to reduce the visual KV write rate while preserving the historical evidence required by subsequent queries.

\subsection{Method Overview}
\label{ssec:method_overview}

EviStream processes each frame in five stages. First, \textbf{Patch-ABD} detects semantic transitions from locally aligned token changes and maintains segment boundaries across streaming calls. Second, \textbf{segment-start and periodic anchor} frames supply a compact causal reference for the current scene. Third, a filter removes \textbf{high-norm artifacts} that would otherwise consume the selection budget. Fourth, a \textbf{Coverage--Novelty Facility Location} objective selects tokens that jointly summarize the current frame and fill gaps in historical semantic coverage. Finally, redundant unselected tokens are merged into their nearest retained representatives, and the resulting k tokens are written to the long-term KV memory.

\paragraph{Causal Coverage Initialization.}
At each segment boundary, we set \(c^{0}_{t,i}=0\), preventing the preceding anchor from suppressing the new context.
For subsequent frames, let \(\hat{x}_{t,i}=x_{t,i}/\|x_{t,i}\|_2\) denote a normalized token.
Historical coverage is defined by the best local match in the latest anchor \(A\):
\begin{equation}
c^{0}_{t,i}
=
\operatorname{clip}_{[0,1]}
\left(
\max_{j\in\mathcal{N}_{3\times3}(i)}
\operatorname{cos}(\hat{x}_{t,i},\hat{a}_{j})
\right).
\label{eq:historical_coverage}
\end{equation}
High coverage indicates content already represented by the recent context, whereas low coverage identifies potentially novel objects, attributes, relations, or states.
Local matching tolerates small spatial displacements.

\paragraph{Unified Selection Objective.}
For a selected set \(S\subseteq U_t\),
\begin{equation}
F_t(S)
=
\sum_{i\in U_t}
\max\!\left(
c^{0}_{t,i},
\max_{j\in S}
\operatorname{cos}(\hat{x}_{t,i},\hat{x}_{t,j})
\right).
\label{eq:facility_objective}
\end{equation}
The objective jointly rewards within-frame representativeness and cross-frame novelty: selected tokens summarize the current frame while emphasizing content not covered by the anchor.
At segment boundaries, the zero coverage baseline reduces the objective to representative scene summarization; within a segment, historical coverage discounts redundant content and emphasizes incremental updates.

\paragraph{Training-free integration.}
EviStream changes neither the visual encoder nor the language model parameters.
It can be inserted as a causal memory layer in an existing streaming VLM and remains compatible with downstream KV-block packing and reasoning.
The default configuration uses \(k=7\), \(p=4\), \(\alpha=15\%\), \(\lambda=5\), and \(\rho=0.8\).

\begin{table*}[t]
\centering
\fontsize{9.2pt}{11.5pt}\selectfont
\setlength{\tabcolsep}{1.8pt}
\setlength{\aboverulesep}{0pt}
\setlength{\belowrulesep}{0pt}
\renewcommand{\arraystretch}{1.1}
\begin{tabular}{cl|cccccccc|c|ccccc|ccc}
\toprule[2.0pt]
 & \textbf{Model}
 & \textbf{CO}${\scriptstyle\uparrow}$ & \textbf{CI}${\scriptstyle\uparrow}$ & \textbf{OD}${\scriptstyle\uparrow}$ & \textbf{PC}${\scriptstyle\uparrow}$
 & \textbf{SC}${\scriptstyle\uparrow}$ & \textbf{TA}${\scriptstyle\uparrow}$ & \textbf{TR}${\scriptstyle\uparrow}$ & \textbf{TJ}${\scriptstyle\uparrow}$
 & \textbf{Acc}${\scriptstyle\uparrow}$
 & \textbf{ER}${\scriptstyle\uparrow}$ & \textbf{CR}${\scriptstyle\uparrow}$ & \textbf{RA}${\scriptstyle\uparrow}$
 & \textbf{CC}${\scriptstyle\uparrow}$ & \textbf{AF}${\scriptstyle\uparrow}$
 & \textbf{Mem}${\scriptstyle\downarrow}$
 & \textbf{TTFT}${\scriptstyle\downarrow}$
 & \textbf{KV}${\scriptstyle\uparrow}$ \\
\midrule
\multirow{6}{*}{\rotatebox{90}{\textbf{Offline}}}
 & Qwen2.5-VL-7B
 & 36.4 & 34.2 & 38.5 & 43.8 & 38.3 & 40.1 & 35.8 & 35.2
 & 38.6
 & 62.7 & 76.4 & 62.5 & 51.4 & 94.3
 & 22.40 & 2.520 & 1.00 \\

 & Qwen3-VL-8B
 & 39.4 & 37.0 & 42.9 & 42.3 & 37.2 & 49.6 & 33.6 & 42.9
 & 40.2
 & 67.3 & 77.2 & 64.8 & 53.4 & 95.7
 & \textbf{19.88} & \textbf{0.757} & \textbf{1.00} \\

 & Qwen3-VL-30B
 & 42.6 & 38.1 & 45.1 & 43.5 & 39.6 & 50.4 & 35.2 & 44.1
 & 41.7
 & 69.2 & 77.8 & 63.2 & 56.3 & 96.7
 & 36.71 & 1.315 & 1.00 \\

 & Qwen2.5-VL-72B
 & 40.6 & 37.7 & 43.6 & 43.7 & 38.1 & 50.6 & 34.3 & 41.7
 & 41.5
 & 68.3 & 78.1 & 65.3 & 55.4 & 95.8
 & 160.37 & 8.306 & 1.00 \\

 & GLM-4.6V
 & 38.8 & 40.1 & 40.4 & 42.3 & 41.3 & 43.4 & 37.4 & 42.3
 & 40.9
 & 69.9 & 79.0 & 69.3 & 55.8 & 95.1
 & 231.33 & 8.793 & 1.00 \\

 & Qwen3-VL-235B
 & \textbf{55.1} & \textbf{49.4} & \textbf{53.0} & \textbf{50.7}
 & \textbf{44.2} & \textbf{57.0} & \textbf{46.3} & \textbf{54.7}
 & \textbf{52.1}
 & \textbf{71.0} & \textbf{80.1} & \textbf{69.8} & \textbf{59.5} & \textbf{97.2}
 & 300.19 & 15.193 & 1.00 \\

\midrule
\multirow{5}{*}{\rotatebox{90}{\textbf{Streaming}}}
 & ReKV~\cite{di2025streaming}
 & 40.9 & 32.2 & 40.3 & 32.6 & 28.7 & 45.0 & 29.1 & 40.4
 & 37.2
 & \textbf{70.8} & 68.0 & 57.4 & 44.2 & 83.2
 & 29.76 & 0.235 & 1.95 \\

 & OASIS~\cite{liang2026oasis}
 & 41.5 & \textbf{40.5} & 35.5 & 37.2 & 32.3 & 43.4 & 28.2 & 43.2
 & 38.1
 & 69.3 & \textbf{79.6} & \textbf{69.5} & \textbf{56.8} & \textbf{91.0}
 & 23.54 & 3.735 & 1.98 \\

 & FluxMem~\cite{xie2026fluxmem}
 & 41.2 & 33.7 & \textbf{42.3} & 34.9 & 31.8 & \textbf{47.7} & 24.2 & 39.6
 & 37.0
 & 70.1 & 69.8 & 60.4 & 46.8 & 84.0
 & 22.43 & 1.971 & 2.07 \\

 & MuKV~\cite{xiao2026mukv}
 & 32.8 & 24.9 & 41.7 & 28.5 & 31.3 & 43.9 & 29.2 & 33.9
 & 34.2
 & 63.6 & 67.3 & 55.4 & 41.5 & 88.1
 & 19.09 & 0.194 & \textbf{35.70} \\

 & \textbf{Ours (EviStream)}
 & \textbf{43.8} & 40.0 & 41.0 & \textbf{43.7} & \textbf{33.9} & 47.4 & \textbf{34.3} & \textbf{48.4}
 & \textbf{43.5}
 & \textbf{72.75} & 77.81 & 68.75 & 55.62 & 88.74
 & \textbf{16.99} & \textbf{0.054} & 22.13 \\
\bottomrule[1.5pt]
\end{tabular}
\caption{Accuracy (\%) of offline and streaming models on StreamRecall across 8 graph reasoning modes:
\textbf{CO} (Concurrency), \textbf{CI} (Cross-Instance), \textbf{OD} (Ordering), \textbf{PC} (Process Chain), \textbf{SC} (State Chain), \textbf{TA} (Temporal Aggregation), \textbf{TR} (Temporal Reasoning), and \textbf{TJ} (Trajectory Tracking); \textbf{Acc} is the overall accuracy.
Think-process metrics (ER, CR, RA, CC, and AF) are averaged over all evaluated questions.
\textbf{Mem} denotes overall peak GPU memory in GiB, \textbf{TTFT} is time-to-first-token, and \textbf{KV} is the physical KV-cache compression ratio ($\times$).
The best result within each group (offline / streaming) is shown in \textbf{bold}.}
\label{tab:reasoning_mode_acc}
\end{table*}

\begin{table}[t]
\centering
\small
\setlength{\tabcolsep}{6pt}
\setlength{\aboverulesep}{0pt}
\setlength{\belowrulesep}{0pt}
\renewcommand{\arraystretch}{1.15}
\begin{tabular}{lcc}
\toprule[1.5pt]
\textbf{Variant}
& \textbf{Cal. Acc. (\%)}
& \(\boldsymbol{\Delta}\) \textbf{(\%)} \\
\midrule
\textbf{EviStream (full)}
& \textbf{43.52} & -- \\
w/o Patch-ABD
& 42.02 & \(-1.50\) \\
w/o Temporal Coverage
& 40.27 & \(-3.25\) \\
w/o Artifact Filter
& 43.27 & \(-0.25\) \\
w/o CovNov Selector
& 41.77 & \(-1.75\) \\
w/o Running-Mean Merge
& 42.02 & \(-1.50\) \\
\bottomrule[1.5pt]
\end{tabular}
\caption{Component ablation of EviStream. \(\Delta\) denotes the absolute accuracy change from the full model.}
\label{tab:module_ablation}
\end{table}

\begin{table}[t]
\centering
\small
\setlength{\tabcolsep}{5pt}
\setlength{\aboverulesep}{0pt}
\setlength{\belowrulesep}{0pt}
\renewcommand{\arraystretch}{1.15}
\begin{tabular}{lcc|cc}
\toprule[1.5pt]
& \multicolumn{2}{c|}{\textbf{LLaVA-OV-7B}}
& \multicolumn{2}{c}{\textbf{Qwen3-VL-8B}} \\
\textbf{Metric}
& \textbf{Native} & \textbf{EviStream}
& \textbf{Native} & \textbf{EviStream} \\
\midrule
Overall (\%) \(\uparrow\)
& 33.83 & \textbf{38.61}
& 41.20 & \textbf{43.52} \\
Mem (GiB) \(\downarrow\)
& 29.42 & \textbf{19.33}
& 19.88 & \textbf{16.99} \\
TTFT (s) \(\downarrow\)
& 2.2600 & \textbf{0.0849}
& 0.7570 & \textbf{0.0540} \\
KV (\(\times\)) \(\uparrow\)
& 1.00 & \textbf{22.13}
& 1.00 & \textbf{22.13} \\
\bottomrule[1.5pt]
\end{tabular}
\caption{Generalization and efficiency of EviStream across VLM backbones. LLaVA-OV-7B denotes LLaVA-OneVision-Qwen2-7B. The better result within each backbone is shown in bold. Consistent with Table \ref{tab:reasoning_mode_acc} metrics.}
\label{tab:backbone_generalization}
\end{table}

\section{Experiments}

\label{sec:experiments}

\subsection{Experimental Dataset}
\label{ssec:experimental_dataset}
We evaluate on StreamingVideoBench using videos from the validation splits of EPIC-KITCHENS-100, Assembly101, and Ego4D. The benchmark provides the questions, ground-truth answers, evidence annotations, query times, and task categories. We consider three answer formats: Single-Choice, Multi-Select, and Open-Ended.

\subsection{Experimental Details}
\label{ssec:experimental_details}

\paragraph{Compared Models.}
We evaluate 6 offline VLMs and five streaming systems as shown in Table ~\ref{tab:reasoning_mode_acc}. Both offline VLMs and streaming systems uniformly sample videos at \(1\,\mathrm{fps}\). For each query, offline VLMs receive at most 128 uniformly sampled frames from \(V[0:t_\text{query}]\), whereas streaming systems sequentially process frames from the complete video.

\subsection{Evaluation Metrics}
\label{ssec:experimental_metrics}

\paragraph{Answer Accuracy.}
Single-Choice and Multi-Select are evaluated using exact-match accuracy, with option order ignored for Multi-Select. Duration predictions are considered correct when their relative error does not exceed \(15\%\). Other Open-Ended responses are rated against the ground-truth answer by a model judge (prompt in Appendix~\ref{app:prompts}) on a \(0\)--\(5\) scale, with scores of at least \(4\) counted as correct.

\subsection{Main Results}
\label{ssec:main_results}

\paragraph{Overall Performance.}
Table~\ref{tab:reasoning_mode_acc} shows that EviStream, built on Qwen3-VL-8B, achieves the highest overall accuracy among streaming methods at \(43.5\%\), outperforming the strongest streaming baseline, OASIS, by \(5.4\) percentage points.
Compared with the offline Qwen3-VL-8B using the same backbone, EviStream improves accuracy by \(3.3\) points.
It also achieves the lowest peak memory (\(16.99\) GiB) and TTFT (\(0.054\) s) among streaming methods with a \(22.13\times\) physical KV-cache compression ratio, providing a favorable accuracy--efficiency trade-off.

\paragraph{Graph-Reasoning Performance.}
EviStream ranks first among streaming methods on 5 of the 8 graph-reasoning modes.
The largest gains occur in Process Chain and Trajectory Tracking, exceeding the strongest baselines by \(6.5\) and \(5.2\) percentage points, respectively, where reasoning depends on causally related evidence distributed across distant moments.
Nevertheless, State Chain and Temporal Reasoning remain challenging across models.

\paragraph{Process-Oriented Evaluation.}
The think-process evaluation further characterizes the evidence-use behavior of EviStream.
It achieves the highest ER among streaming methods (\(72.75\)), indicating stronger coverage of multi-hop evidence, while remaining competitive in citation correctness and role assignment.
Its CC of \(55.62\) and AF of \(88.74\) further show that the retained evidence supports largely complete and faithful reasoning chains.

\paragraph{Robustness and Backbone Generalization.}
Tables~\ref{tab:module_ablation} and~\ref{tab:backbone_generalization} show that the improvements are not specific to a particular backbone.
Removing Temporal Coverage causes the largest accuracy drop (\(3.25\) points), confirming the importance of context-aware evidence retention.
EviStream improves accuracy by \(4.78\) points on LLaVA-OV-7B and \(2.32\) points on Qwen3-VL-8B while consistently reducing peak memory and TTFT.
These results demonstrate the generality of EviStream as a training-free streaming memory layer.

\section{Conclusion}

We present StreamRecall, a contextually grounded benchmark for multi-hop streaming video QA, and the Evidence Utilization Cascade for diagnosing evidence-reasoning failures. We further introduce EviStream, a training-free causal memory that preserves uncovered semantic evidence under a fixed KV budget.
Experiments demonstrate state-of-the-art streaming performance, improved efficiency, and consistent gains across VLM backbones.

\bibliography{aaai2027}


\clearpage
\input{files/appendix}

\end{document}

%% file: files/appendix.tex
\appendix

\input{files/appendix_overview}

\input{files/streamrecall}
\input{files/evistream}
\input{files/experiments}
\input{files/ablation}
\input{files/prompts}
\input{files/data_ethics}

%% file: files/appendix_overview.tex
\noindent\textbf{Appendix Overview}
\vspace{0.3em}

\begingroup
\footnotesize
\setlength{\emergencystretch}{1em}
\newcommand{\appoverviewentry}[2]{#1\nobreak\dotfill\nobreak\mbox{\pageref{#2}}}
\noindent
\begin{itemize}[leftmargin=1.5em, itemsep=1pt, parsep=0pt, topsep=2pt]
    \item \appoverviewentry{\textbf{Appendix A}: StreamRecall}{app:streamrecall}
    \begin{itemize}[leftmargin=1.2em, itemsep=0pt, parsep=0pt, topsep=0pt]
        \item \appoverviewentry{A.1~~Source Dataset Background}{app:source_datasets}
        \item \appoverviewentry{A.2~~Detailed Data Construction}{app:data_construction_details}
        \item \appoverviewentry{A.3~~Human Annotation and Quality Control}{app:human_annotation}
        \item \appoverviewentry{A.4~~Detailed Evaluation Metrics}{app:detailed_metrics}
    \end{itemize}

    \item \appoverviewentry{\textbf{Appendix B}: EviStream}{app:evistream}
    \begin{itemize}[leftmargin=1.2em, itemsep=0pt, parsep=0pt, topsep=0pt]
        \item \appoverviewentry{B.1~~Per-Frame Causal Update}{app:causal_update}
        \item \appoverviewentry{B.2~~Causal Semantic Boundary Detection}{app:causal_boundary}
        \item \appoverviewentry{B.3~~Anchor State and Candidate Filtering}{app:anchor_filter}
        \item \appoverviewentry{B.4~~Greedy Coverage--Novelty Optimization}{app:greedy_covnov}
        \item \appoverviewentry{B.5~~Redundancy-Aware Token Merging}{app:token_merging}
        \item \appoverviewentry{B.6~~Streaming Integration and Complexity}{app:streaming_integration}
    \end{itemize}

    \item \appoverviewentry{\textbf{Appendix C}: Experiments}{app:experiments}
    \begin{itemize}[leftmargin=1.2em, itemsep=0pt, parsep=0pt, topsep=0pt]
        \item \appoverviewentry{C.1~~Additional Results}{app:additional_results}
        \item \appoverviewentry{C.2~~Evidence Attribution Probe}{app:evidence_attribution}
    \end{itemize}

    \item \appoverviewentry{\textbf{Appendix D}: Ablation Study}{app:ablation}
    \begin{itemize}[leftmargin=1.2em, itemsep=0pt, parsep=0pt, topsep=0pt]
        \item \appoverviewentry{D.1~~Visual-KV Budget}{app:budget_ablation}
        \item \appoverviewentry{D.2~~Selection Strategy}{app:selection_ablation}
        \item \appoverviewentry{D.3~~Hyperparameter Sensitivity}{app:scalar_sensitivity}
        \item \appoverviewentry{D.4~~Frame-Rate Sensitivity}{app:fps_sensitivity}
    \end{itemize}

    \item \appoverviewentry{\textbf{Appendix E}: Prompts}{app:prompts}
    \item \appoverviewentry{\textbf{Appendix F}: Data Availability, Licensing, and Ethics}{app:data_ethics}
\end{itemize}

\endgroup

\vspace{0.8em}
\noindent\rule{\linewidth}{0.4pt}
\vspace{0.5em}

%% file: files/streamrecall.tex
\section{StreamRecall}


\label{app:streamrecall}

\subsection{Source Dataset Background}
\label{app:source_datasets}

StreamRecall draws source videos from the official validation splits of three established activity datasets.
Together, they cover both first- and third-person viewpoints and span unscripted daily activities and structured procedures.

\paragraph{EPIC-KITCHENS-100.}
EPIC-KITCHENS-100~\cite{damen2022epickitchens} contains 100 hours of unscripted egocentric video recorded in 45 kitchens.
Its 700 long-form videos are densely annotated with approximately 90K action segments, including verb and noun labels for fine-grained hand--object interactions.
The dataset provides realistic object use, state changes, and spatial transitions in everyday kitchen activities.

\paragraph{Assembly101.}
Assembly101~\cite{sener2022assembly101} records participants assembling and disassembling 101 toy vehicles without fixed instructions.
It contains 4,321 videos totaling 513 hours, captured simultaneously from eight static third-person cameras and four egocentric cameras.
Its fine- and coarse-grained action annotations cover ordering variations, mistakes, corrections, and intermediate assembly states, supporting long-horizon procedural reasoning across viewpoints.

\paragraph{Ego4D.}
Ego4D~\cite{grauman2022ego4d} is a large-scale egocentric dataset containing more than 3,000 hours of in-the-wild daily activities collected across diverse participants, locations, and scenarios.
Its broad coverage of household, workplace, outdoor, and leisure activities complements the procedure-focused content of EPIC-KITCHENS-100 and Assembly101 and increases the contextual diversity of StreamRecall.

\subsection{Detailed Data Construction}
\label{app:data_construction_details}

\paragraph{Source Alignment and Temporal Segmentation.}
We first map the source annotations of each dataset to a unified event schema containing the action, object, and absolute temporal boundaries.
Each video is then divided into clips of at most 10 seconds, while all annotations retain their original video timestamps.
The short clips expose local details that are difficult to recover from a long video in one pass, whereas the absolute timestamps preserve cross-clip continuity.
For example, an annotated ``put'' event can be refined into placing a particular bowl on the counter beside the cutting board without losing its position in the full activity timeline.

\paragraph{Anchored VLM Enrichment.}
As shown in Figure~\ref{fig:prompt_enrichment}, the VLM receives each clip together with its structured annotations, a canonical object vocabulary, the preceding object history, and the current activity state.
Rather than recognizing actions from scratch, it enriches the supplied events with visible identity, state, source and destination, tool use, and manipulation details.
The canonical vocabulary constrains entity names across clips, while the history prevents a state change from being interpreted as a new object.
For instance, a generic annotation such as ``remove wheel'' can be enriched with the wheel position, the tool used for removal, its destination on the table, and the resulting assembly state.

\paragraph{Instance Disambiguation.}
Repeated object mentions are grouped using the prompt in Figure~\ref{fig:prompt_disambiguation}.
The LLM first examines activity continuity, including whether an object is picked up, moved, placed, and later reused, and then treats visual attributes as supporting evidence.
Mentions are split only when the timeline and a stable attribute, such as functional role or assembly position, jointly indicate different physical instances.
Thus, a screwdriver observed during both disassembly and reassembly remains one instance despite changes in viewpoint, whereas front-left and rear-right wheels form separate trajectories.
State transitions such as raw-to-cooked or detached-to-attached never create new instances.

\paragraph{Event Merging and Object-Centric Chains.}
Events are ordered within each resolved object trajectory.
Temporally adjacent events with the same action and instance are merged, and their complementary spatial, tool, and state details are consolidated into one caption.
Merging stops at an object change, action transition, or meaningful state boundary.
For example, several consecutive clips showing continuous rinsing are summarized as one interval, while the subsequent placement of the rinsed object remains a separate event.
This step converts fragmented clip descriptions into compact object-centric action chains while preserving temporal order.

\paragraph{QA Generation.}
Using the prompts in Figure~\ref{fig:prompt_qa_generation}, an LLM first interleaves the object-centric chains into a chronological narrative.
It then proposes questions from the 24 task categories by identifying object journeys, state transitions, temporal relations, repeated tool use, concurrency, and process dependencies.
Questions are written without timestamps and must reflect an information need that is meaningful in the current activity.
For each proposal, a VLM retrieves relevant events from the complete evidence pool, after which an LLM derives the answer using only the retrieved timeline.
For example, when a person is about to reuse a knife on vegetables, the pipeline may ask what the knife previously contacted; answering requires linking the earlier cutting, cleaning, and reuse events.

\paragraph{Automatic Quality Verification.}
The verification prompt in Figure~\ref{fig:prompt_verification} checks the proposed answer against both the retrieved captions and the corresponding video clips.
It identifies supporting, contradictory, and missing evidence and rejects candidates whose answers are unsupported or whose evidence chain is incomplete.
As an additional sufficiency test, a VLM independently answers the question using only the designated evidence clips.
A candidate proceeds to human review only when the evidence-only answer agrees with the proposed ground truth.

\paragraph{Trigger Point Assignment.}
The trigger process in Figure~\ref{fig:prompt_trigger} first restricts candidate query points to events occurring after the end of the latest required evidence.
A VLM identifies later events that could activate the information need, and an LLM selects the earliest contextually appropriate candidate.
In the knife example, the trigger is not the earlier cutting event that supplies the answer, but the later moment when the person reaches for the knife again and its prior use becomes relevant.
This separates the time at which evidence is observed from the time at which the question naturally arises.

\subsection{Human Annotation and Quality Control}
\label{app:human_annotation}

\paragraph{Annotator Roles and Calibration.}
Human verification follows a primary-review, independent-verification, and adjudication protocol.
Each candidate is first reviewed by one annotator and independently checked by a second annotator who cannot see the first decision.
A senior annotator resolves disagreements and audits a sample of both accepted and rejected candidates.
Before full annotation, all annotators complete a shared calibration batch and discuss disagreements using a common guideline covering contextual relevance, answer correctness, evidence sufficiency, and reasoning-pattern assignment.
Videos and source datasets are rotated across annotators to reduce annotator-specific and domain-specific bias.

\paragraph{Context-Aware QA Screening.}
Annotators watch the video prefix up to \(t_{\mathrm{query}}\) and assess whether the question is motivated by the scene at the assigned trigger.
A question is rejected when it is merely answerable from the annotations but does not correspond to a plausible information need in the current activity.
For example, asking about the prior use of a knife when it is about to be reused is retained, whereas asking the same question at an unrelated scene transition is removed.
Annotators also reject questions that expose future information, contain timestamps, or can be answered solely from a single current frame when multi-event recall is intended.

\paragraph{Answer Verification and Correction.}
Without consulting the generated answer, annotators first derive an answer directly from the source video and then compare it with the proposed ground truth.
They correct entity references, counts, durations, orderings, locations, and state descriptions when the proposal is semantically incomplete or inconsistent with the video.
The revised answer must match the question scope exactly; supporting facts that are not requested are excluded.
Candidates are removed when the video does not support a unique answer.

\paragraph{Evidence Completeness and Uniqueness.}
Annotators verify every cited interval against the source video and revise its temporal boundary when necessary.
The final evidence set must satisfy three criteria: each item contributes to the answer, the set jointly contains all required facts, and the question can be answered from these clips without access to the remaining video.
We retain only questions for which reviewers agree on one canonical evidence set.
Candidates admitting multiple disjoint sufficient evidence sets, or requiring information outside the designated clips, are removed.
This protocol ensures a one-to-one mapping from each QA pair to a complete evidence set rather than multiple alternative annotations.

\paragraph{Agreement and Adjudication.}
The two reviewers independently record accept/reject decisions, answer edits, evidence edits, query-time validity, scene labels, and graph-reasoning labels.
Agreement is assessed separately for these fields so that a high-level acceptance decision cannot conceal evidence or label disagreements.
Any disagreement affecting the answer or evidence chain is resolved by replaying the relevant clips during adjudication; uncertain cases are removed rather than settled by majority preference.
Guidelines are updated only between annotation rounds and then applied retrospectively to the affected batch, preventing annotator-specific criteria from drifting over time.

\paragraph{Effect of Human Verification.}
Before human review, each candidate contains an automatically generated question, answer, evidence set, trigger point, and task labels, and may exhibit contextual mismatch, answer errors, incomplete evidence, or ambiguous support.
Human review either removes the candidate or converts it into a context-aware QA pair with a corrected answer, one complete evidence set, a causally valid query time, and verified scene and reasoning labels.
The resulting benchmark contains 4,983 QA pairs: 1,537 from EPIC-KITCHENS-100, 1,401 from Assembly101, and 2,045 from Ego4D, together grounded in 14,968 evidence items.

\subsection{Detailed Evaluation Metrics}
\label{app:detailed_metrics}

Let \(\mathcal{Q}_{\mathrm{think}}\) denote the set of \(N\) samples with valid
reasoning outputs and valid judge responses. VD, HalR, and EPR are computed per
sample and then averaged over \(\mathcal{Q}_{\mathrm{think}}\), whereas CTP and
FEP are aggregated directly from the sample-level stage indicators defined
below.

\paragraph{Visual Dependency and Hallucination Rate.}
For each sample \(q\), the judge assigns \(v_q\in\{0,1\}\), where \(v_q=1\)
indicates that the specific answer requires video observation and cannot be
derived from the question, answer options, common sense, or world knowledge
alone. Let \(n^{\mathrm{claim}}_q\) be the number of specific claims about the
video in the reasoning output and \(n^{\mathrm{hal}}_q\) the number unsupported
by the gold evidence. We compute
\begin{align}
\mathrm{VD}
&=
\frac{1}{N}\sum_{q\in\mathcal{Q}_{\mathrm{think}}}v_q,
\label{eq:vd}\\
\mathrm{HalR}
&=
\frac{1}{N}\sum_{q\in\mathcal{Q}_{\mathrm{think}}}
\frac{n^{\mathrm{hal}}_q}
{\max(1,n^{\mathrm{claim}}_q)}.
\label{eq:halr}
\end{align}
Thus, VD measures the average dependence on visual evidence, whereas HalR
measures the mean fraction of unsupported video claims. Both metrics lie in
\([0,1]\); higher VD and lower HalR are preferable.

\paragraph{Cascade Pass Indicators.}
Let the ordered stages be
\(\mathcal{S}=(\mathrm{Locate},\mathrm{Cite},\mathrm{Ground},
\mathrm{Complete},\mathrm{Derive})\), and let \(z_{q,k}\in\{0,1\}\) indicate
whether sample \(q\) passes stage \(k\). Locate passes when
\(\mathrm{HR}_q=1\); Cite passes when \(\mathrm{CR}_q=1\) and at least one hop
is located; Ground passes when \(\mathrm{RA}_q=1\) and at least one hop is
cited; Complete passes when \(\mathrm{CC}_q\geq0.8\); and Derive passes when
\(\mathrm{AF}_q=1\). These strict indicators support the following
stage-level diagnostics.

\paragraph{Cascade Transition Probability.}
For adjacent stages \(k\) and \(k+1\), CTP is the conditional probability of
passing the next stage after passing the current stage:
\begin{equation}
\mathrm{CTP}_{k\rightarrow k+1}
=
\frac{
\sum_{q\in\mathcal{Q}_{\mathrm{think}}}z_{q,k}z_{q,k+1}
}{
\sum_{q\in\mathcal{Q}_{\mathrm{think}}}z_{q,k}
}.
\label{eq:ctp}
\end{equation}
The metric is reported when the denominator is nonzero. A low transition
probability identifies a bottleneck between the corresponding stages.

\paragraph{First Error Position.}
For each sample, let
\(f_q=\min\{k:z_{q,k}=0\}\); if all five stages pass, we set
\(f_q=\mathrm{NoError}\). For
\(k\in\mathcal{S}\cup\{\mathrm{NoError}\}\), the FEP distribution is
\begin{equation}
\mathrm{FEP}(k)
=N^{-1}\!\sum_{q\in\mathcal{Q}_{\mathrm{think}}}
\mathbf{1}[f_q=k].
\label{eq:fep}
\end{equation}
FEP therefore localizes the earliest failure in the cascade rather than
counting all downstream errors caused by that failure.

\paragraph{Error Propagation Rate.}
Let \(E_q\) count all failed stages for sample \(q\), and let \(P_q\) count
failed stages whose immediately preceding stage also fails:
\begin{equation}
\begin{aligned}
E_q &= \sum_{k=1}^{5}(1-z_{q,k}),\\
P_q &= \sum_{k=2}^{5}(1-z_{q,k-1})(1-z_{q,k}).
\end{aligned}
\end{equation}
The reported EPR averages the fraction of propagated failures per sample:
\begin{equation}
\mathrm{EPR}
=
\frac{1}{N}
\sum_{q\in\mathcal{Q}_{\mathrm{think}}}
\frac{P_q}{\max(1,E_q)}.
\label{eq:epr}
\end{equation}
A sample with no failed stage has EPR \(0\). Higher values indicate that stage
failures tend to persist downstream, whereas lower values indicate isolated
failures or subsequent recovery.

%% file: files/evistream.tex
\section{EviStream}
\label{app:evistream}

This section provides the implementation details omitted from
the main text of the paper, including causal boundary detection, state
maintenance, greedy token selection, and backbone integration.

\subsection{Per-Frame Causal Update}
\label{app:causal_update}

Let \(Z_{t-1}=(P_{t-1},\mathcal{H}_{t-1},u_{t-1},A_{t-1})\) denote the
selector state before frame \(t\), comprising the preceding-frame tokens,
boundary-score history, within-segment counter, and latest full anchor.
EviStream performs the causal update
\begin{equation}
\begin{aligned}
(S_t,Z_t)&=\Phi(X_t,Z_{t-1}), \qquad |S_t|=k,\\
\widetilde{X}_t&=\operatorname{Merge}(X_t,S_t),\\
\mathcal{K}_t&=\operatorname{Update}
(\mathcal{K}_{t-1},\widetilde{X}_t).
\end{aligned}
\label{eq:app_causal_update}
\end{equation}
Here, \(\Phi\) applies Patch-ABD, updates the anchor state, filters invalid
candidates, and optimizes semantic coverage. Neither a future frame nor an
unrevealed question enters this update. Questions are processed only after
the visual state at their query time has been established.

\subsection{Causal Semantic Boundary Detection}
\label{app:causal_boundary}

\paragraph{Locally Aligned Transition Score.}
Patch-ABD estimates semantic change from the complete set of \(N\) projected
visual tokens, independently of the subsequent KV budget. To tolerate small
object displacements, each normalized token in the current frame is matched
only within the corresponding \(3\times3\) neighborhood of the preceding
frame. The local change and frame-level transition score are
\begin{equation}
\begin{aligned}
d_{t,i}
&=1-\max_{j\in\mathcal{N}_{3\times3}(i)}
\operatorname{cos}(\hat{x}_{t,i},\hat{x}_{t-1,j}),\\
r_t
&=\operatorname{mean}\!\left(
\operatorname{Top}_{\lceil\alpha N\rceil}(d_{t,:})
\right),
\end{aligned}
\label{eq:app_transition_score}
\end{equation}
where \(\alpha=0.15\), corresponding to 30 of the 196 patches. Local
matching absorbs motion within one patch while preventing unrelated regions
from explaining each other.

\paragraph{Adaptive Causal Decision.}
We causally smooth the score as
\(\widetilde{r}_t=0.75r_t+0.25r_{t-1}\). Let
\(\mathcal{H}_{t-1}\) contain at most 120 preceding smoothed transition
scores. The boundary threshold is
\begin{equation}
\begin{aligned}
\theta_t
&=\operatorname{median}(\mathcal{H}_{t-1})\\
&\quad+2.5\max\!\left(
1.4826\operatorname{MAD}(\mathcal{H}_{t-1}),10^{-4}
\right).
\end{aligned}
\label{eq:app_boundary_threshold}
\end{equation}
During the first three transitions, we use the warm-up threshold
\(2.5\times10^{-4}\). A transition is accepted as a boundary only when its
smoothed score meets or exceeds \(\theta_t\), rises above the preceding score, and
satisfies the minimum segment length and inter-boundary distance of two
frames. Thus, all decisions depend exclusively on observations available by
time \(t\).

\paragraph{Persistent Boundary State.}
The preceding-frame tokens, transition history, latest smoothed score,
segment counter, and anchor state persist across streaming calls. An API call
boundary is therefore not treated as a semantic boundary: processing the
same frames in one call or in consecutive chunks produces identical
frame-level decisions. Only an accepted Patch-ABD boundary resets the segment
counter and starts a new anchor state.

\subsection{Anchor State and Candidate Filtering}
\label{app:anchor_filter}

\paragraph{Anchor Update.}
The first frame of each detected segment becomes a segment anchor, and the
anchor is refreshed every \(p=4\) frames within the segment. The selector
retains all normalized tokens of only the latest anchor as short-term state;
this representation is never written in full to the language-model KV cache.
For a periodic anchor frame, token selection is first conditioned on the
preceding anchor, after which the full current representation replaces it.
Consequently, an anchor refresh does not erase the novelty of the current
frame before selection.

\paragraph{High-Norm Artifact Filtering.}
Register-like visual tokens with unusually large norms can repeatedly consume
the selection budget. For \(n_i=\lVert x_{t,i}\rVert_2\), we define
\begin{equation}
\gamma_t=
\operatorname{median}(n)
+\lambda\,1.4826
\max\!\left(\operatorname{MAD}(n),10^{-6}\right),
\label{eq:app_artifact_threshold}
\end{equation}
with \(\lambda=5\). Tokens above \(\gamma_t\) are excluded from both the
candidate and coverage sets. We remove at most eight tokens per frame and
always preserve at least \(k\) valid candidates, yielding the set \(U_t\)
used in Eq.~\ref{eq:facility_objective}.

\subsection{Greedy Coverage--Novelty Optimization}
\label{app:greedy_covnov}

Equation~\ref{eq:facility_objective} is optimized under the cardinality
constraint \(|S_t|=k\). Starting with \(S_t=\varnothing\) and
\(c_i=c^0_{t,i}\), each iteration selects the candidate with the largest
marginal coverage gain and updates the covered similarity:
\begin{equation}
\begin{aligned}
j^{\star}
&=\arg\max_{j\in U_t\setminus S_t}
\sum_{i\in U_t}\max\!\left(
0,\operatorname{cos}(\hat{x}_{t,i},\hat{x}_{t,j})-c_i
\right),\\
c_i
&\leftarrow\max\!\left(
c_i,\operatorname{cos}(\hat{x}_{t,i},\hat{x}_{t,j^{\star}})
\right).
\end{aligned}
\label{eq:app_greedy_covnov}
\end{equation}
We then set \(S_t\leftarrow S_t\cup\{j^{\star}\}\) and repeat until
\(|S_t|=k\). Once a semantic region is represented, selecting a similar token
provides little additional gain; the fixed budget therefore spreads across
uncovered objects and regions rather than concentrating on one salient area.
Selected tokens are packed in their original spatial order, preserving their
positions independently of the greedy selection order.

\subsection{Redundancy-Aware Token Merging}
\label{app:token_merging}

Hard selection would discard every token outside \(S_t\). Instead, each
unselected token is assigned to its most similar retained representative. Let
\(m_j\) be the number of tokens already absorbed by representative
\(\bar{x}_{t,j}\), initialized with \(m_j=1\) and
\(\bar{x}_{t,j}=x_{t,j}\). The receiver and running-mean update are
\begin{equation}
\begin{aligned}
j_r^{\star}
&=\arg\max_{j\in S_t}
\operatorname{cos}(x_{t,r},x_{t,j}),\\
\bar{x}_{t,j_r^{\star}}
&\leftarrow
\frac{m_{j_r^{\star}}\bar{x}_{t,j_r^{\star}}+x_{t,r}}
{m_{j_r^{\star}}+1},\\[-2pt]
&\hspace{-2em}\text{if }\operatorname{cos}
(x_{t,r},x_{t,j_r^{\star}})\geq\rho.
\end{aligned}
\label{eq:app_token_merging}
\end{equation}
where \(\rho=0.8\); \(m_{j_r^{\star}}\) is incremented after each accepted
merge. Tokens below the threshold are discarded rather than forced into an
unrelated representative. Merging thus preserves cluster-level information
without increasing the number of KV entries.

\subsection{Streaming Integration and Complexity}
\label{app:streaming_integration}

\paragraph{Backbone Integration.}
For the Qwen3-VL implementation, retained tokens preserve the original
three-dimensional M-RoPE coordinates \((t,h,w)\) on the \(14\times14\)
visual grid rather than being re-indexed after compression. A separate
contiguous physical cache-position sequence is used for KV storage. Each
frame retains the native block structure
\(\texttt{vision\_start}\rightarrow k\text{ selected patches}
\rightarrow\texttt{vision\_end}\). The same retained positions and merge
assignments are applied to the main visual stream and all DeepStack feature
streams, ensuring exact token alignment across layers.

\paragraph{Multi-Round State Management.}
Video frames update the persistent visual KV cache and selector state before
the content or arrival time of a future question is known. At a query time,
question and decoding tokens are appended temporarily for answer generation.
They are removed after the answer, while the visual KV cache and selector
state are retained, allowing ingestion to resume without making later visual
compression dependent on earlier questions. Future frames are never used,
and discarded visual tokens are not reconstructed.

\paragraph{Memory and Computation.}
Without compression, \(T\) frames write \(NT\) visual patch tokens; EviStream
writes \(kT\), giving a visual-patch write ratio of \(k/N\). The latest full
anchor requires \(O(Nd)\) short-term selector state but does not enter the
language-model KV cache. With a precomputed pairwise similarity matrix, the
selector requires \(O(N^2d+kN^2)\) operations and \(O(N^2)\) temporary memory
per frame. It does not reduce visual-encoder computation because all \(N\)
tokens are produced before selection.

\begin{table}[t]
\centering
\small
\setlength{\tabcolsep}{5pt}
\renewcommand{\arraystretch}{1.08}
\caption{Default EviStream configuration.}
\label{tab:app_evistream_hyperparameters}
\begin{tabular}{lc}
\toprule
\textbf{Parameter} & \textbf{Value} \\
\midrule
Visual tokens per frame \(N\) & 196 \\
KV patch budget \(k\) & 7 \\
Patch-ABD neighborhood & \(3\times3\) \\
Transition aggregation ratio \(\alpha\) & 0.15 \\
Boundary history length & 120 \\
Boundary MAD multiplier & 2.5 \\
Anchor refresh period \(p\) & 4 frames \\
Artifact multiplier \(\lambda\) & 5 \\
Maximum filtered tokens & 8 \\
Merge threshold \(\rho\) & 0.8 \\
\bottomrule
\end{tabular}
\end{table}


%% file: files/experiments.tex
\section{Experiments}
\label{app:experiments}

\subsection{Additional Results}
\label{app:additional_results}

\subsubsection{Task-Level Accuracy}
\label{app:task_type_results}

\begin{figure*}[t]
\centering
\includegraphics[width=\textwidth]{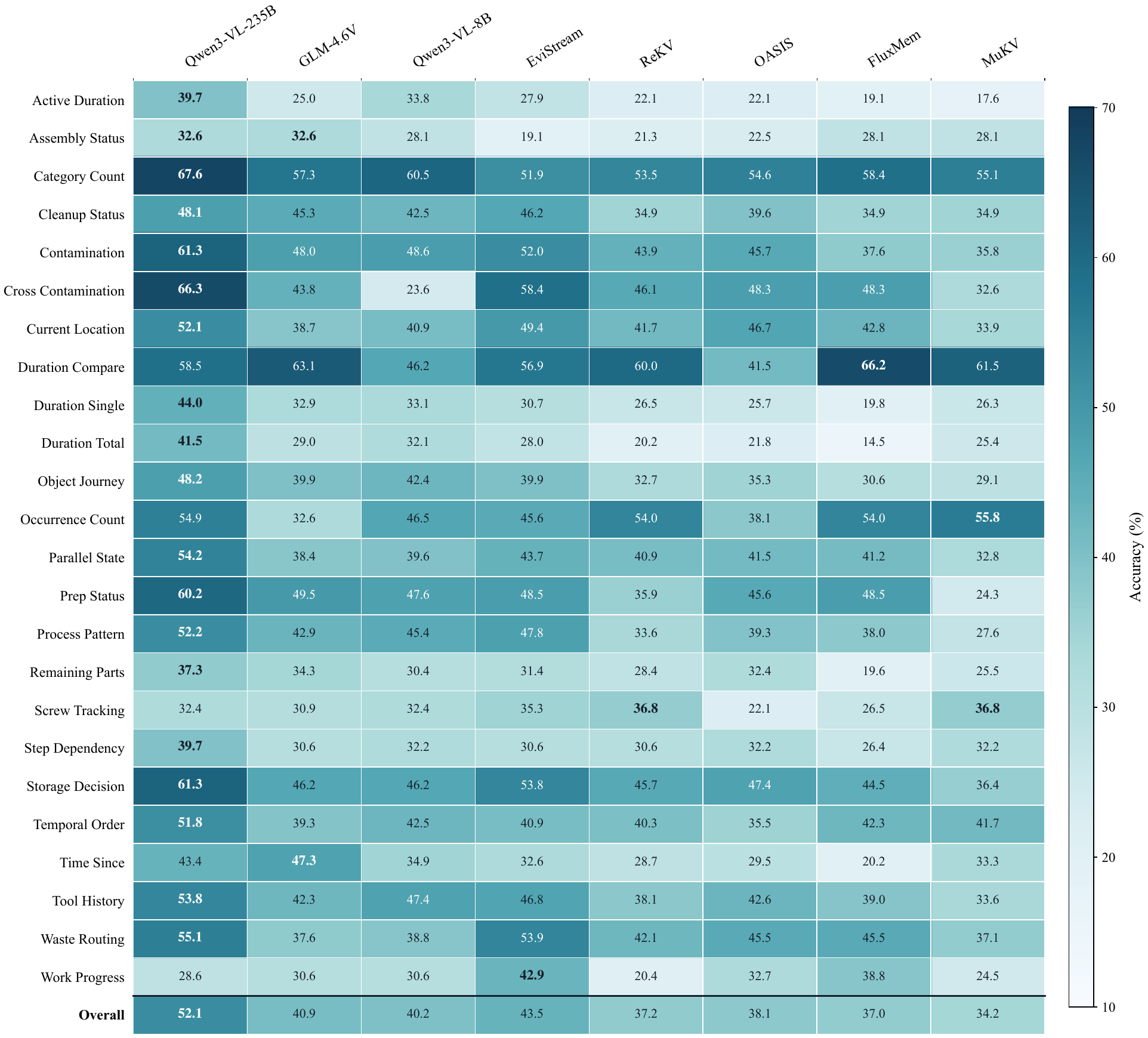}
\caption{No-think accuracy across the 24 scenario-grounded QA categories. The
first three columns correspond to offline VLMs, while EviStream and the
remaining four columns correspond to streaming systems. Bold values indicate
the highest accuracy in each row.}
\label{fig:nonthink_accuracy_by_type}
\end{figure*}

\paragraph{Task-Level Performance.}
Figure~\ref{fig:nonthink_accuracy_by_type} shows that EviStream achieves the
highest overall accuracy among streaming systems at \(43.5\%\), outperforming
the strongest streaming baseline, OASIS, by \(5.4\) percentage points.
Qwen3-VL-235B remains the strongest offline model at \(52.1\%\), leaving an
\(8.6\)-point gap between EviStream and the largest offline VLM.

Across task categories, EviStream ranks first among streaming systems on 14 of
24 categories and ties for first on Prep Status. Its largest margins over the
strongest streaming baseline for each category occur on Cross Contamination
(\(+10.1\) points), Process Pattern (\(+8.5\)), and Waste Routing
(\(+8.4\)), demonstrating stronger retention of temporally distributed object
states and interactions.

\begin{figure}[t]
  \centering
  \includegraphics[width=\linewidth]{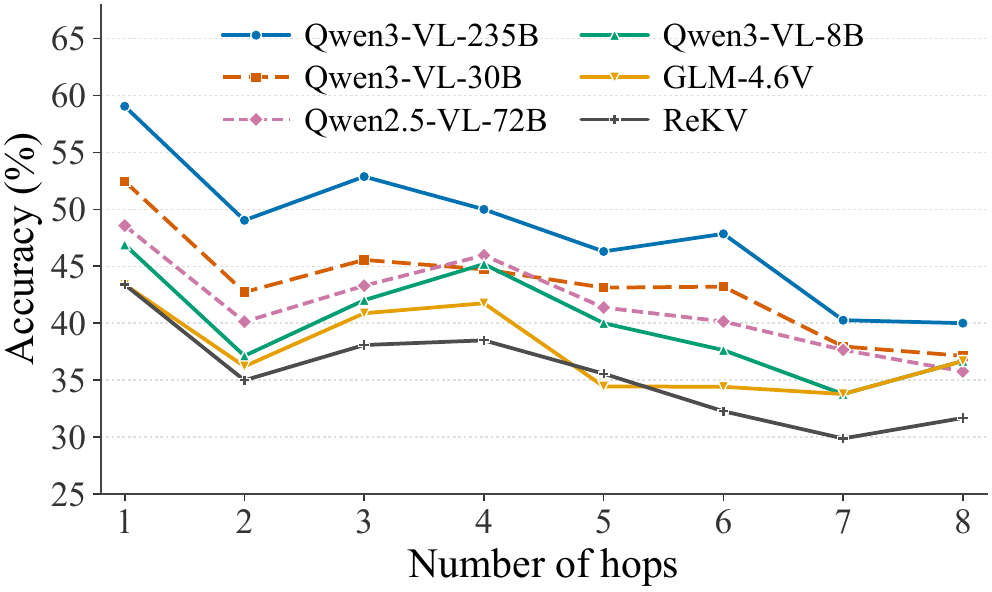}
  \caption{No-think answer accuracy as a function of the number of annotated
  evidence hops. Each hop corresponds to a distinct evidence item required to
  answer the question.}
  \label{fig:accuracy_by_hop_count}
\end{figure}

\begin{figure}[t]
  \centering
  \includegraphics[width=\linewidth]{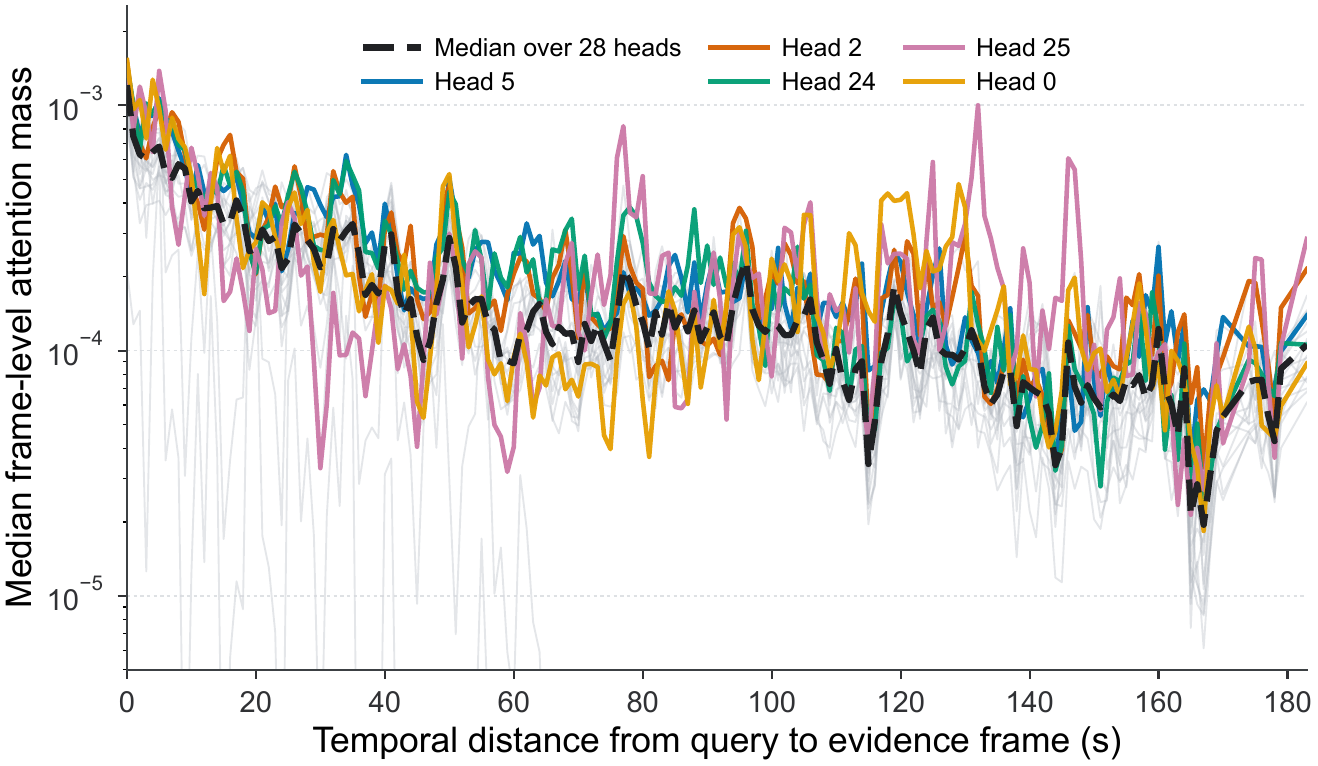}
  \caption{Layer-27 attention probe for LLaVA-OneVision-7B. At the generation
  position of the key answer word, we measure the attention assigned by each of
  28 heads to evidence-frame patches whose intermediate representations assign
  high vocabulary probability to evidence-related words. Light gray curves
  show individual heads, colored curves highlight representative heads, and
  the dashed black curve denotes the median over all heads.}
  \label{fig:attention_distance_probe}
\end{figure}

\subsubsection{Hop Count and Temporal Scope}

\paragraph{Hop Count and Evidence Scope.}
Figure~\ref{fig:accuracy_by_hop_count} shows a general decline in accuracy as
the number of required evidence hops increases, despite local non-monotonic
variation. This trend indicates that integrating more evidence imposes
additional reasoning pressure on both offline VLMs and streaming systems.
However, hop count is not an isolated measure of reasoning complexity:
questions with more evidence items often cover a longer temporal interval.
The observed degradation may therefore reflect both the combinatorial burden
of integrating multiple observations and the difficulty of retaining evidence
over an expanded temporal scope.

\paragraph{Effect of Temporal Evidence Distance.}
To examine the temporal component directly, we probe attention as a function
of \(\Delta t=t_{\mathrm{query}}-t_{\mathrm{evidence}}\).
Figure~\ref{fig:attention_distance_probe} shows that the median attention mass
assigned to evidence-related patches decreases overall as \(\Delta t\) grows.
Although individual heads exhibit occasional spikes, the aggregate trend shows
a clear decline on the logarithmic scale. Thus, evidence farther from the query
has progressively weaker influence on answer generation, supporting the need
for a streaming memory that preserves and exposes temporally distant evidence
for downstream reasoning.

\subsection{Evidence Attribution Probe}
\label{app:evidence_attribution}

\paragraph{Controlled Setup.}
We compare CovNov-7 with Random-7 and Uniform-7 under the same EviStream
configuration. All variants use Qwen3-VL-8B, process video causally at
\(1\,\mathrm{fps}\), and retain \(k=7\) of 196 patches per frame. Patch-ABD,
anchor updates, artifact filtering, running-mean merging with threshold
\(0.8\), positional encoding, prompting, and greedy decoding are held fixed;
only the seven retained locations differ. For normalized patch features,
CovNov-7 selects the set \(S\) that greedily maximizes
\begin{equation}
F(S)=
\sum_{i=1}^{196}
\max\!\left(c_i^0,\max_{j\in S}\cos(x_i,x_j)\right),
\label{eq:attr_covnov}
\end{equation}
where \(c_i^0\) measures the coverage supplied by the historical anchor.
Random-7 samples seven valid locations with a deterministic seed, whereas
Uniform-7 uses seven fixed locations distributed across the \(14\times14\)
grid. The probe covers 33 Open-Ended
questions with complete human evidence annotations. Comparisons are matched by
dataset, question type, hop count, and query--evidence distance, and annotated
evidence intervals are extended by two seconds on each side without crossing
the query time.

\paragraph{Evidence-Related Patches.}
For each QA, an LLM extracts at most four evidence-related words \(w\) that
appear verbatim in the evidence captions and are informative for deriving the
answer. From the Full-196 replay, we apply the language-model output head to the
Layer-26 patch state \(h^{26}_{t,i}\), immediately before Layer 27, and define
\begin{equation}
\begin{aligned}
p_{t,i,w} &= P\!\left(w\mid h^{26}_{t,i}\right),\\
H_{t,i,w} &= \mathbf{1}\!\left[p_{t,i,w}>10^{-4}\right].
\end{aligned}
\label{eq:attr_high_patch}
\end{equation}
Thus, \(H_{t,i,w}\) identifies patches whose intermediate representations
assign high probability to evidence-related word \(w\).

\paragraph{Stage I: Direct Evidence Selection.}
Let \(S_t\) denote the seven locations selected in evidence frame \(t\), and
let \(\mathcal{E}\) be the set of annotated evidence frames. We collect the
evidence-related patch indices as
\(\Omega_w=\{(t,i):t\in\mathcal{E},H_{t,i,w}=1\}\). Direct recall is
\begin{equation}
R_{\mathrm{direct},w}=
\frac{\left|\{(t,i)\in\Omega_w:i\in S_t\}\right|}
{\left|\Omega_w\right|}.
\label{eq:attr_direct_recall}
\end{equation}
This metric measures how often the seven retained locations directly capture
evidence-related patches.

\paragraph{Stage II: Evidence Retention after Merging.}
For each unselected patch \(x_{t,i}\), the closest selected representative is
\begin{equation}
\begin{aligned}
j^* &= \arg\max_{j\in S_t}\cos(x_{t,i},x_{t,j}),\\
\cos(x_{t,i},x_{t,j^*}) &\geq 0.8.
\end{aligned}
\label{eq:attr_merge_receiver}
\end{equation}
A patch satisfying the threshold contributes to the representative through
running-mean merging. Let \(N_{\mathrm{high},w}\) be the number of
evidence-related patches, and let \(N_{\mathrm{direct},w}\) and
\(N_{\mathrm{merged},w}\) count those retained directly and through merging.
The post-merge recall is
\begin{equation}
R_{\mathrm{retained},w}=
\frac{N_{\mathrm{direct},w}+N_{\mathrm{merged},w}}
{N_{\mathrm{high},w}}.
\label{eq:attr_retained_recall}
\end{equation}

\paragraph{Stage III: Evidence Utilization during Generation.}
We extract the Layer-27 query by teacher-forcing answer-bearing words from the
ground-truth answer. To control for the number and identity of retained keys,
queries from all methods attend to the same Full-196 key bank containing only
patches within the annotated evidence intervals. The attention share assigned
to evidence-related patches is
\begin{equation}
A_{\mathrm{high}}=
\frac{\sum_{(t,i):H_{t,i,w}=1}
A(q_{\mathrm{target}},k_{t,i})}
{\sum_{(t,i)\in\mathcal{E}}
A(q_{\mathrm{target}},k_{t,i})},
\label{eq:attr_attention_share}
\end{equation}
and the controlled attention change is
\begin{equation}
\Delta A_{\mathrm{method}}=
A_{\mathrm{method}}-A_{\mathrm{Full\text{-}196}}.
\label{eq:attr_attention_delta}
\end{equation}
All metrics are first averaged within each QA. Paired differences and 95\%
confidence intervals are estimated with 5,000 QA-level bootstrap samples.

\begin{table}[t]
\centering
\footnotesize
\setlength{\tabcolsep}{3pt}
\renewcommand{\arraystretch}{1.12}
\begin{tabular}{lccc}
\toprule
\textbf{Metric} & \textbf{CovNov} & \textbf{Random} & \textbf{Uniform} \\
\midrule
Direct recall (\%) & \textbf{4.02} & 3.20 & 1.86 \\
Retained recall (\%) & \textbf{9.05} & 5.99 & 4.74 \\
Merge gain (\%) & \textbf{5.03} & 2.79 & 2.88 \\
L27 attention $\Delta$ & \textbf{0.000795} & 0.000131 & 0.000016 \\
\bottomrule
\end{tabular}
\caption{Evidence attribution under an identical seven-token visual KV budget.
The attention change is measured relative to Full-196.}
\label{tab:app_evidence_attribution}
\end{table}

\paragraph{Results.}
Table~\ref{tab:app_evidence_attribution} shows that CovNov provides the highest
direct recall and that its main advantage emerges after merging. Retained
evidence recall exceeds Random-7 by \(3.06\) points (95\% CI:
\([0.79,5.60]\)) and Uniform-7 by \(4.31\) points (95\% CI:
\([1.47,6.98]\)). CovNov also yields a larger controlled Layer-27 attention
change than Random-7, with a paired difference of \(0.000664\) (95\% CI:
\([0.000052,0.001525]\)). These results show that CovNov selects effective
representative receivers, enabling running-mean merging to preserve more
evidence within the fixed KV budget. The resulting memory further strengthens
the routing of answer-generation queries toward relevant evidence.


%% file: files/ablation.tex
\section{Ablation Study}
\label{app:ablation}

\paragraph{Scope and Protocol.}
The main paper reports the final (k=7) component ablation in
Table~\ref{tab:module_ablation} and the cross-backbone generalization and
efficiency results in Table~\ref{tab:backbone_generalization}. This appendix
provides complementary analyses of the KV budget, selection strategy,
hyperparameters, and frame rate.
Unless otherwise specified, all ablations use a fixed 400-question subset
uniformly sampled from the full evaluation set, comprising 196
Single-Choice (SC), 69 Multi-Select (MC), and 135 Open-Ended (OE) questions.
Every comparison within a table uses the same questions, backbone, sampling
pipeline, and answer evaluator, and all OE predictions receive valid
judgments. The reported values are raw subset accuracies and should be
compared only within each controlled experiment, rather than with the
full-set calibrated results in the main paper.

\subsection{Visual-KV Budget}
\label{app:budget_ablation}

We first fix the complete Coverage--Novelty selector and vary only the number
of visual patches written to the long-term KV cache per frame.
Table~\ref{tab:app_budget_sweep} shows that the highest subset accuracy is
obtained with \(k=56\), but this setting writes eight times as many patches as
\(k=7\). The adopted \(k=7\) configuration retains only
\(7/196=3.57\%\) of the original patches while remaining within 1.75
percentage points of the highest-accuracy setting. We therefore use it as the
efficiency-oriented operating point in the main experiments.

\begin{table}[t]
\centering
\scriptsize
\setlength{\tabcolsep}{3.5pt}
\setlength{\aboverulesep}{0pt}
\setlength{\belowrulesep}{0pt}
\renewcommand{\arraystretch}{1.08}
\caption{Accuracy under different per-frame visual-KV budgets. KV write is
relative to the adopted \(k=7\) configuration.}
\label{tab:app_budget_sweep}
\begin{tabular}{rccc}
\toprule[1.5pt]
\(\boldsymbol{k}\) & \textbf{OE (\%)} & \textbf{Overall (\%)} &
\textbf{Rel. KV} \\
\midrule
5  & 18.52 & 39.75 & \(0.71\times\) \\
6  & 19.26 & 40.50 & \(0.86\times\) \\
7  & 20.00 & 41.00 & \(1\times\) \\
14 & 20.00 & 39.50 & \(2\times\) \\
28 & 19.26 & 40.75 & \(4\times\) \\
42 & 20.00 & 42.50 & \(6\times\) \\
56 & 22.22 & \textbf{42.75} & \(8\times\) \\
\bottomrule[1.5pt]
\end{tabular}
\end{table}

\subsection{Selection Strategy}
\label{app:selection_ablation}

Table~\ref{tab:app_selection_strategy} isolates the choice of the seven
retained positions. Patch-ABD, anchor updates, artifact filtering,
running-mean merging, frame sampling, and the backbone remain unchanged.
Coverage--Novelty exceeds Random and Uniform selection by 1.75 and 2.75
percentage points, respectively. Uniform selection uses a staggered
\(2\text{--}3\text{--}2\) layout over the \(14\times14\) grid and replaces
an artifact-filtered target with the nearest valid patch. Its lower accuracy
indicates that spatial coverage alone is insufficient under an extreme
budget; selection must also account for frame content and historical
coverage.

\begin{table}[t]
\centering
\scriptsize
\setlength{\tabcolsep}{2.0pt}
\setlength{\aboverulesep}{0pt}
\setlength{\belowrulesep}{0pt}
\renewcommand{\arraystretch}{1.08}
\caption{Controlled comparison of selection strategies at \(k=7\).
All accuracy values are percentages; \(\Delta\) is relative to
Coverage--Novelty.}
\label{tab:app_selection_strategy}
\begin{tabular}{lccccc}
\toprule[1.5pt]
\textbf{Strategy} & \textbf{SC} & \textbf{MC} & \textbf{OE} &
\textbf{Overall (\%)} & \(\boldsymbol{\Delta}\) \\
\midrule
CovNov-7  & 65.82 & 14.49 & 20.00 & \textbf{41.00} & -- \\
Random-7  & 65.31 &  8.70 & 17.04 & 39.25 & \(-1.75\) \\
Uniform-7 & 61.22 & 10.14 & 19.26 & 38.25 & \(-2.75\) \\
\bottomrule[1.5pt]
\end{tabular}
\end{table}

\subsection{Hyperparameter Sensitivity}
\label{app:scalar_sensitivity}

\begin{figure*}[t]
\centering
\includegraphics[width=\textwidth]{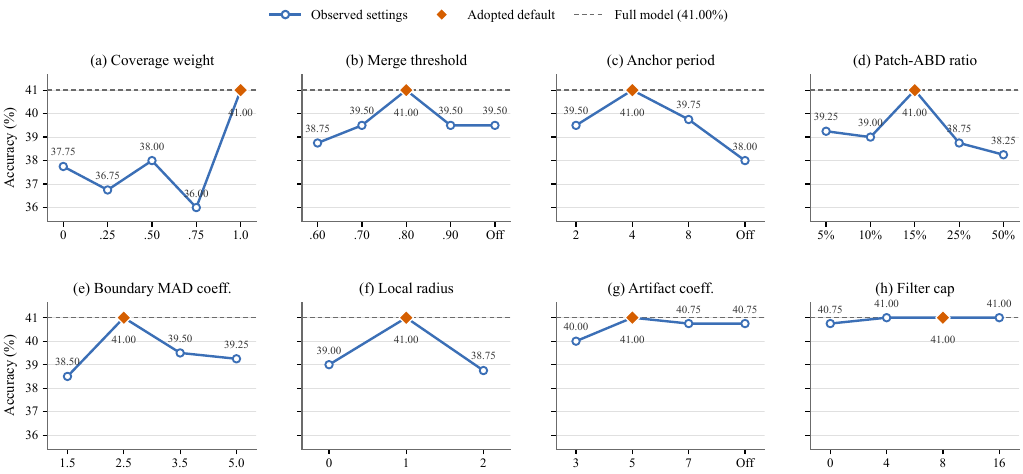}
\caption{One-variable hyperparameter sensitivity at \(k=7\) on the
400-question subset. Blue circles denote evaluated settings, orange diamonds
mark the adopted defaults, and the dashed line indicates the full-model
accuracy of \(41.00\%\). All other settings are held fixed within each panel.}
\label{tab:app_scalar_sensitivity}
\end{figure*}

Figure~\ref{tab:app_scalar_sensitivity} varies one scalar at a time around the
default \(k=7\) configuration. No tested alternative exceeds the default
accuracy of 41.00\%. In particular, discounting anchor-derived coverage
consistently degrades performance, while disabling periodic anchors reduces
accuracy by 3.00 points. The selected merge threshold, Patch-ABD aggregation
ratio, boundary threshold, and local radius are also the best values within
their respective sweeps. Artifact filtering is less sensitive, with all
tested variants lying within one point of the default.

\subsection{Frame-Rate Sensitivity}
\label{app:fps_sensitivity}

The frame-rate sweep in Table~\ref{tab:app_fps_sweep} uses the complete
method with \(k=7\), varying only the causal
sampling rate. Performance varies non-monotonically with sampling density:
very sparse sampling yields the lowest accuracy, while performance peaks at
1.2~fps before declining at 1.3~fps. This result indicates that adequate
temporal coverage is necessary, but denser sampling does not guarantee further
gains under a fixed per-frame token budget. We therefore retain 1~fps as the
default, since increasing the rate to 1.2~fps improves overall accuracy by only
0.20 percentage points.

\begin{table}[t]
\centering
\scriptsize
\setlength{\tabcolsep}{2.4pt}
\setlength{\aboverulesep}{0pt}
\setlength{\belowrulesep}{0pt}
\renewcommand{\arraystretch}{1.05}
\caption{Frame-rate sensitivity with \(k=7\). Accuracy and \(\Delta\) are
reported in percentage points, with \(\Delta\) measured relative to 1~fps.}
\label{tab:app_fps_sweep}
\begin{tabular}{rccccc}
\toprule[1.5pt]
\textbf{fps} & \textbf{SC} & \textbf{MC} & \textbf{OE} &
\textbf{Overall (\%)} & \(\boldsymbol{\Delta}\) \\
\midrule
0.1 & 62.76 & 11.42 & 13.33 & 35.13 & \(-5.87\) \\
0.2 & 63.27 & 12.31 & 17.04 & 36.36 & \(-4.64\) \\
0.4 & 64.80 & 12.86 & 17.04 & 37.17 & \(-3.83\) \\
0.6 & 65.82 & 12.17 & 18.52 & 37.05 & \(-3.95\) \\
0.8 & 64.80 & 13.52 & 19.32 & 39.54 & \(-1.46\) \\
1.0 & 65.82 & 14.49 & 20.00 & 41.00 & \(0.00\) \\
1.1 & 65.75 & 15.10 & 19.55 & 41.10 & \(+0.10\) \\
1.2 & 65.93 & 14.49 & 20.06 & \textbf{41.20} & \(+0.20\) \\
1.3 & 66.18 & 13.04 & 18.25 & 40.18 & \(-0.82\) \\
\bottomrule[1.5pt]
\end{tabular}
\end{table}

%% file: files/prompts.tex
\section{Prompts}
\label{app:prompts}
%

\definecolor{dcPromptBlue}{HTML}{D6E6FF}
\definecolor{dcPromptMint}{HTML}{D7F9F8}
\definecolor{dcPromptCream}{HTML}{FFFFEA}
\definecolor{dcPromptPeach}{HTML}{FFF0D4}
\definecolor{dcPromptPink}{HTML}{FBE0E0}
\definecolor{dcPromptLavender}{HTML}{E5D4EF}
\definecolor{dcPromptLilac}{HTML}{F2EAF7}
\definecolor{dcPromptNearWhite}{HTML}{F9F5FB}

\begin{figure*}[t]
\centering
\begin{tcolorbox}[
    colback=dcPromptBlue,
    colframe=dcPromptMint,
    colbacktitle=dcPromptMint,
    coltitle=black,
    title={\textbf{Prompt for Anchored VLM Enrichment}},
    fonttitle=\small,
    fontupper=\small,
    width=\textwidth,
    boxrule=0.8pt,
    arc=3pt,
]
\raggedright
You are a precise activity-annotation assistant. You will watch a short
\(\leq 10\)-second clip from a \texttt{\{viewpoint\}} video of
\texttt{\{activity\_domain\}}.

\smallskip
\noindent\textbf{Inputs}

\noindent\textbf{Ground-truth action annotations:}
\texttt{\{segment\_annotations\}}\\
\textbf{Canonical object vocabulary:}
\texttt{\{canonical\_objects\}}\\
\textbf{Object history before this clip:}
\texttt{\{object\_history\}}\\
\textbf{Current activity progress:}
\texttt{\{activity\_progress\}}

\medskip
\noindent\textbf{Task}

The action identities and temporal boundaries are already provided. Do
\textbf{not} replace them with unconstrained action recognition. Watch the clip
and enrich every annotated event with the following missing information:

\smallskip
\noindent
\textbf{1.\ Object identity and state.}
Resolve generic mentions using the canonical vocabulary and prior object
history. Record appearance, orientation, preparation/assembly state, and other
visual attributes only when they help distinguish the object or track its
state over time.\\
\textbf{2.\ Spatial context.}
Specify the source location, destination location, supporting surface,
container, attachment point, and salient object-to-object relations.\\
\textbf{3.\ Tool or means.}
Verify the annotated tool against the visible clip. If no tool is annotated
but one is visibly used, record the observed tool or manipulation method.\\
\textbf{4.\ Activity progress.}
Describe the state transition caused by the event and how it changes the
current cooking, assembly, work, or household process.\\
\textbf{5.\ Manipulation details.}
When visible and useful, record the acting hand, grip, motion direction,
rotation direction, repetition count, and other process details.

\medskip
\noindent\textbf{Critical Rules}

\begin{enumerate}
    \item The \texttt{object} field must exactly match one entry in
    \texttt{\{canonical\_objects\}}; do not invent synonyms or new entities.
    \item Produce one enriched event per supplied annotation. If an annotated
    event extends beyond the clip, describe only the visible portion.
    \item Add an extra event only when a clearly visible action is missing from
    the structured annotations and it involves a canonical object.
    \item Preserve temporal continuity with \texttt{\{object\_history\}} and
    \texttt{\{activity\_progress\}}. Do not turn a state change into a new
    physical object.
    \item Report only visually supported details. Omit trivial attributes,
    unsupported intent, and speculative causal explanations.
    \item Use \texttt{\{perspective\_expression\}} consistently, e.g.,
    first person for egocentric videos and ``the person'' for exocentric videos.
\end{enumerate}

\smallskip
\noindent\textbf{Output Format (JSON only)}

\noindent\texttt{\{}\\
\texttt{\quad "events": [\{}\\
\texttt{\qquad "timestamp": "\{relative\_time\}",}\\
\texttt{\qquad "action": "\{anchored\_action\}",}\\
\texttt{\qquad "object": "\{canonical\_object\}",}\\
\texttt{\qquad "detail": "\{state, tool, and manipulation details\}",}\\
\texttt{\qquad "source": "\{specific source location\}",}\\
\texttt{\qquad "destination": "\{specific destination location\}",}\\
\texttt{\qquad "count": \{integer\}}\\
\texttt{\quad \}],}\\
\texttt{\quad "spatial\_relations": ["\{visible relations\}"],}\\
\texttt{\quad "end\_snapshot": "\{state and layout at clip end\}"}\\
\texttt{\}}
\end{tcolorbox}
\caption{
    Unified anchored-enrichment prompt distilled from the dataset-specific
    EPIC-KITCHENS, Assembly101, and Ego4D scripts. Structured annotations and a
    canonical object vocabulary constrain the VLM, while video evidence supplies
    state, spatial, tool, progress, and manipulation details.
}
\label{fig:prompt_enrichment}
\end{figure*}

\begin{figure*}[t]
\centering
\begin{tcolorbox}[
    colback=dcPromptCream,
    colframe=dcPromptPeach,
    colbacktitle=dcPromptPeach,
    coltitle=black,
    title={\textbf{Prompt for Context-Aware Instance Disambiguation}},
    fonttitle=\small,
    fontupper=\small,
    width=\textwidth,
    boxrule=0.8pt,
    arc=3pt,
]
\raggedright
You are resolving whether repeated mentions of one object name in a long
activity video refer to the same physical instance or to multiple distinct
instances.

\smallskip
\noindent\textbf{Inputs}

\noindent
\textbf{Target object name:} \texttt{\{object\_name\}}\\
\textbf{All target-object events in chronological order:}
\texttt{\{indexed\_object\_events\}}\\
\textbf{Context before, during, and after the target events:}
\texttt{\{surrounding\_events\}}

\medskip
\noindent\textbf{Reasoning Procedure}

\noindent\textbf{Step 1 --- Context First.}
Read the full activity timeline as a coherent workflow. Determine whether the
person is continuously interacting with one item, whether the item is moved and
later revisited, and whether multiple instances are plausible in this activity.

\smallskip
\noindent\textbf{Step 2 --- Visual Descriptions Second.}
Compare color, size, position, orientation, and other descriptions. Treat such
differences as weak evidence because viewpoint, lighting, partial visibility,
and VLM wording may vary across clips.

\smallskip
\noindent\textbf{Split only when both conditions hold:}

\begin{enumerate}
    \item the temporal context shows separate handling episodes for different
    physical items; and
    \item the items have a stable inherent distinction, such as persistent
    color, size, functional role, or assembly position.
\end{enumerate}

\smallskip
\noindent\textbf{Do not split when:}

\begin{itemize}
    \item one item is picked up, used, placed, and later reused;
    \item the apparent difference is a state change, e.g., raw to cooked,
    loose to attached, or partial to completed;
    \item a tool is repeatedly picked up and put down;
    \item descriptions plausibly differ because of viewpoint or annotation noise;
    \item identical fasteners or ingredients are treated collectively and no
    stable instance identity is required.
\end{itemize}

\smallskip
\noindent\textbf{Naming Rules}

Use concise, lower-case, identity-based names such as
\texttt{front wheel} or \texttt{red pepper}. Never use arbitrary numbered names
such as \texttt{wheel 1}. Assign every input event index exactly once.

\medskip
\noindent\textbf{Output Format (JSON only)}

\noindent If all mentions refer to one object:\\
\texttt{\{"split": false, "reason": "\{brief explanation\}"\}}

\smallskip
\noindent If multiple instances are supported:\\
\texttt{\{"split": true, "reason": "\{brief explanation\}",}\\
\texttt{\quad "instances": [}\\
\texttt{\qquad \{"name": "\{specific name\}", "event\_indices": [\{indices\}]\}}\\
\texttt{\quad ]\}}
\end{tcolorbox}
\caption{
    Prompt for resolving repeated object mentions into temporally consistent
    physical trajectories. Activity continuity is evaluated before visual
    description differences, reducing false splits caused by viewpoint changes
    and inconsistent captions.
}
\label{fig:prompt_disambiguation}
\end{figure*}

\begin{figure*}[t]
\centering
\begin{tcolorbox}[
    colback=dcPromptLilac,
    colframe=dcPromptLavender,
    colbacktitle=dcPromptLavender,
    coltitle=black,
    title={\textbf{Prompts for Multi-Hop QA Generation}},
    fonttitle=\small,
    fontupper=\scriptsize,
    width=\textwidth,
    boxrule=0.8pt,
    arc=3pt,
]
\raggedright
\begin{minipage}[t]{0.485\textwidth}
\textbf{A.\ Chronological Narrative Prompt}

\smallskip
You receive per-object event chains from a
\texttt{\{viewpoint\}} \texttt{\{activity\_domain\}} video. Each event contains
a timestamp, action, object, and enriched caption.

\smallskip
\noindent\textbf{Task:} Write a complete chronological narrative from the first
event to the last. Interleave events across objects by timestamp; track object
state changes, locations, tool reuse, simultaneous activities, process progress,
and transitions between activity phases. Cover every meaningful event concisely
and finish with a summary of the session and its duration.

\smallskip
\noindent\textbf{Input:}
\texttt{\{per-object timestamped event chains\}}\\
\textbf{Output:} \texttt{\{chronological scene narrative\}}

\medskip
\hrule
\medskip
\textbf{B.\ Question-Proposal Prompt}

\smallskip
Read the narrative as a coherent workflow. Identify natural temporal
connections, including object journeys, repeated tool use, state transitions,
concurrent activities, prerequisites, and unresolved safety or progress needs.
Then propose questions that a real observer would naturally ask at a meaningful
moment in the activity.

\smallskip
\noindent\textbf{Candidate QA Types (24)}

\begin{tabular}{@{}p{0.28\linewidth}p{0.67\linewidth}@{}}
\textbf{Counting/progress} &
\texttt{category\_count}, \texttt{occurrence\_count},
\texttt{work\_progress}\\
\textbf{Temporal} &
\texttt{duration\_single}, \texttt{duration\_compare},
\texttt{duration\_total}, \texttt{temporal\_order},
\texttt{time\_since}, \texttt{active\_duration}\\
\textbf{Tracking} &
\texttt{tool\_history}, \texttt{current\_location},
\texttt{object\_journey}\\
\textbf{Process/state} &
\texttt{process\_pattern}, \texttt{cleanup\_status},
\texttt{parallel\_state}\\
\textbf{Resource flow} &
\texttt{prep\_status}, \texttt{waste\_routing},
\texttt{storage\_decision}\\
\textbf{Safety} &
\texttt{contamination}, \texttt{cross\_contamination}\\
\textbf{Assembly} &
\texttt{assembly\_status}, \texttt{remaining\_parts},
\texttt{screw\_tracking}, \texttt{step\_dependency}\\
\end{tabular}

\smallskip
\noindent\textbf{Critical Rules}

\begin{enumerate}
    \item Each question must require at least two distinct earlier evidence
    moments; reject questions answerable from one event or one current frame.
    \item The question must express a realistic information need for the
    activity domain, not an arbitrary query generated only because annotations
    are available.
    \item Do not place explicit timestamps in the question. Timestamps may
    appear only in evidence and trigger fields.
    \item Answers must be concrete and verifiable: a count, duration, order,
    object set, location, state, process, or causal prerequisite.
    \item Use the dataset-appropriate perspective and exact canonical object names.
    \item Avoid duplicates and distribute questions across applicable QA types.
\end{enumerate}

\smallskip
\noindent\textbf{Output (JSON array)}

\texttt{[\{"question":"\{question\}",}\\
\texttt{\quad "type":"\{one of 24 types\}",}\\
\texttt{\quad "trigger":"\{natural information-need scenario\}",}\\
\texttt{\quad "trigger\_time":"\{approximate candidate time\}",}\\
\texttt{\quad "relevant\_objects":["\{canonical objects\}"]\}]}
\end{minipage}
\hfill
\begin{minipage}[t]{0.485\textwidth}
\textbf{C.\ Semantic Evidence-Retrieval Prompt}

\smallskip
You receive one question and one batch of object event logs. Select every object
whose events may help answer the question.

\smallskip
\noindent An object is relevant when:

\begin{itemize}
    \item it is explicitly mentioned or directly queried;
    \item its events mention an action, location, state, tool, or object needed
    by the question;
    \item it provides necessary temporal or causal context;
    \item it belongs to a semantic category requested by the question; or
    \item its caption describes an interaction involving another queried object.
\end{itemize}

Be inclusive: retaining a borderline-relevant object is preferable to omitting
a necessary evidence chain. Use exact input object names.

\smallskip
\noindent\textbf{Input}

\textbf{Question:} \texttt{\{question\}}\\
\textbf{Object-event batch:} \texttt{\{timestamped event logs\}}

\smallskip
\noindent\textbf{Output (JSON only)}

\texttt{\{"relevant\_objects":["\{objects\}"],}\\
\texttt{\quad "reasoning":"\{brief relevance explanation\}"\}}

\medskip
\hrule
\medskip
\textbf{D.\ Evidence-Grounded Answer Prompt}

\smallskip
You receive the question, QA type, proposed trigger scenario, a unified
chronological timeline of all retrieved events, and limited surrounding context.

\smallskip
\noindent\textbf{Reasoning Procedure}

\begin{enumerate}
    \item Read the full retrieved timeline from start to end and reason across
    objects. An event recorded under one object may imply an interaction with
    another object.
    \item Base the final answer only on retrieved evidence. Surrounding events
    may clarify the workflow but must not introduce unsupported answer facts.
    \item For counts, scan all retrieved events and remove duplicate descriptions
    of the same activity-level occurrence.
    \item For durations, identify relevant start/end events or sum the specified
    active intervals. For locations, use the latest supported placement before
    the query point. For order and dependency, preserve temporal direction.
    \item Return the minimal set of distinct evidence hops that jointly entails
    the answer.
\end{enumerate}

\smallskip
\noindent\textbf{Input}

\textbf{Question:} \texttt{\{question\}}\\
\textbf{Type:} \texttt{\{qa\_type\}}\\
\textbf{Trigger scenario:} \texttt{\{trigger\}}\\
\textbf{Retrieved timeline:} \texttt{\{retrieved\_events\}}\\
\textbf{Surrounding context:} \texttt{\{context\_events\}}

\smallskip
\noindent\textbf{Output (JSON only)}

\texttt{\{"answer":"\{concrete answer\}",}\\
\texttt{\quad "reasoning":"\{brief evidence-grounded derivation\}",}\\
\texttt{\quad "evidence":[}\\
\texttt{\qquad \{"time":"\{time\}","object":"\{object\}",}\\
\texttt{\qquad\quad "event":"\{supporting event\}"\}],}\\
\texttt{\quad "hop\_count":\{number of distinct hops\},}\\
\texttt{\quad "confidence":"high|medium|low"\}}
\end{minipage}
\end{tcolorbox}
\caption{
    Four-prompt QA-construction sequence: chronological narration, realistic
    multi-hop question proposal, batched semantic evidence retrieval, and
    evidence-grounded answer generation. The listed 24 QA types are the
    deduplicated union of the domain-specific categories used across
    EPIC-KITCHENS, Assembly101, and Ego4D.
}
\label{fig:prompt_qa_generation}
\end{figure*}

\begin{figure*}[t]
\centering
\begin{tcolorbox}[
    colback=dcPromptPink,
    colframe=dcPromptNearWhite,
    colbacktitle=dcPromptNearWhite,
    coltitle=black,
    title={\textbf{Prompts for Answer Verification and Evidence Sufficiency}},
    fonttitle=\small,
    fontupper=\scriptsize,
    width=\textwidth,
    boxrule=0.8pt,
    arc=3pt,
]
\raggedright
\begin{minipage}[t]{0.485\textwidth}
\textbf{A.\ Caption-Level Answer Verification}

\smallskip
You are verifying a proposed QA pair against chronological event captions.

\smallskip
\noindent\textbf{Inputs}

\textbf{Question:} \texttt{\{question\}}\\
\textbf{Proposed answer:} \texttt{\{answer\}}\\
\textbf{Proposed reasoning:} \texttt{\{reasoning\}}\\
\textbf{Cited evidence:} \texttt{\{cited\_events\}}\\
\textbf{Event-caption batch:} \texttt{\{caption\_batch\}}

\smallskip
\noindent\textbf{Task}

Read every caption in the batch and determine whether it:

\begin{itemize}
    \item \textbf{supports} a claim in the proposed answer;
    \item \textbf{contradicts} the answer or shows a later state/location;
    \item reveals \textbf{missing} information required for completeness; or
    \item contains no evidence relevant to the question.
\end{itemize}

Check exact counts, sequence direction, duration comparisons, latest known
locations, complete object sets, state transitions, and process dependencies.
Evidence distributed across different objects or batches must be considered
jointly before the QA pair is accepted.

\smallskip
\noindent\textbf{Output (JSON only)}

\texttt{\{"supporting":["\{supported claims\}"],}\\
\texttt{\quad "contradicting":["\{contradictions\}"],}\\
\texttt{\quad "missing":["\{omitted required facts\}"],}\\
\texttt{\quad "batch\_verdict":"correct|incomplete|wrong|no\_evidence"\}}

\medskip
\noindent\textbf{Final caption-level decision:}
Accept only if all relevant batches jointly support a correct and complete
answer, with no unresolved contradiction or missing required fact.
\end{minipage}
\hfill
\begin{minipage}[t]{0.485\textwidth}
\textbf{B.\ Evidence-Only Video Sufficiency Check}

\smallskip
You are given only the video segments cited as gold evidence, not the full
video. Each segment is expanded slightly around its evidence timestamp, and
overlapping intervals are merged.

\smallskip
\noindent\textbf{Input}

\textbf{Evidence clips:} \texttt{\{gold\_evidence\_clips\}}\\
\textbf{Question:} \texttt{\{question\}}\\
\textbf{Options, when applicable:} \texttt{\{options\}}

\smallskip
\noindent\textbf{Task}

Answer the question using only the supplied evidence clips:

\begin{itemize}
    \item for single choice, output one option letter;
    \item for multi-select, output all correct letters;
    \item for open-ended questions, output one direct, concise sentence;
    \item do not use external assumptions or information from uncited video
    intervals.
\end{itemize}

\smallskip
\noindent For duration questions, first caption each evidence clip with its
timestamp, actions, objects, tools, and state/location changes. Then answer from
the ordered timestamped captions, computing or summing time intervals as needed.

\smallskip
\noindent\textbf{Evidence-Sufficiency Criterion}

The cited evidence is sufficient only if an independent model can recover the
gold answer from the evidence-only input. For ordered answers the sequence must
match; for counts the value must match; for objects, states, locations, and
processes the key facts must be semantically equivalent; for durations the
derived value must satisfy the predefined numerical tolerance.

\smallskip
\noindent\textbf{Recorded Output}

\texttt{\{"predicted\_answer":"\{answer from evidence only\}",}\\
\texttt{\quad "answer\_match":"correct|partial|incorrect",}\\
\texttt{\quad "evidence\_sufficient":true|false\}}

\medskip
\noindent\textbf{Reject} QA pairs that are wrong, incomplete, contradicted,
unsupported by the cited clips, dependent on unavailable future evidence, or
based on unreliable fine-grained distinctions.
\end{minipage}
\end{tcolorbox}
\caption{
    Two-level quality-verification prompt set. Caption-level verification searches
    all event logs for support, contradiction, and omissions, while an independent
    evidence-only re-answering test checks whether the cited video segments jointly
    suffice to recover the gold answer.
}
\label{fig:prompt_verification}
\end{figure*}

\begin{figure*}[t]
\centering
\begin{tcolorbox}[
    colback=dcPromptNearWhite,
    colframe=dcPromptBlue,
    colbacktitle=dcPromptCream,
    coltitle=black,
    title={\textbf{Prompts for Two-Stage Trigger Point Assignment}},
    fonttitle=\small,
    fontupper=\small,
    width=\textwidth,
    boxrule=0.8pt,
    arc=3pt,
]
\raggedright
\begin{minipage}[t]{0.485\textwidth}
\textbf{Stage 1 --- VLM Candidate Trigger Selection}

\smallskip
You are selecting candidate trigger events for a streaming-video QA benchmark.
A trigger is a moment when a real viewer would naturally need the queried
information. It must not occur before all evidence required to answer the
question has appeared.

\smallskip
\noindent\textbf{Inputs}

\textbf{Question:} \texttt{\{question\}}\\
\textbf{Answer:} \texttt{\{answer\}}\\
\textbf{Evidence timestamps:} \texttt{\{evidence\_times\}}\\
\textbf{Earliest allowed time:}
\(\max_i t^{\mathrm{end}}_{\mathrm{evidence},i}\)\\
\textbf{Chronological event batch:} \texttt{\{event\_batch\}}

\smallskip
\noindent\textbf{Task}

Select events at or after the earliest allowed time that could naturally
activate the information need. Good candidates include:

\begin{itemize}
    \item the start of a related action;
    \item reuse of a previously observed object or tool;
    \item a transition to a new activity phase;
    \item a safety-, progress-, or memory-relevant decision point;
    \item completion of a meaningful milestone.
\end{itemize}

Do not select an event merely because it is temporally valid. Explain why the
question becomes useful at that moment.

\smallskip
\noindent\textbf{Output (JSON only)}

\texttt{\{"candidates":[}\\
\texttt{\quad \{"time":"\{timestamp\}",}\\
\texttt{\quad\quad "event":"\{trigger event\}",}\\
\texttt{\quad\quad "why":"\{natural information need\}"\}]\}}

\smallskip
If no event qualifies, output
\texttt{\{"candidates":[]\}}.
\end{minipage}
\hfill
\begin{minipage}[t]{0.485\textwidth}
\textbf{Stage 2 --- LLM Final Trigger Selection}

\smallskip
You are assigning the final query timestamp \(t_{\mathrm{query}}\). You receive
the QA pair, its evidence timestamps, and candidate trigger events that have
already passed the causal constraint.

\smallskip
\noindent\textbf{Inputs}

\textbf{Question:} \texttt{\{question\}}\\
\textbf{Answer:} \texttt{\{answer\}}\\
\textbf{Evidence timestamps:} \texttt{\{evidence\_times\}}\\
\textbf{Earliest allowed time:}
\texttt{\{latest evidence end\}}\\
\textbf{Candidate events:} \texttt{\{candidate\_triggers\}}

\smallskip
\noindent\textbf{Selection Rules}

\begin{enumerate}
    \item Reject any candidate earlier than the end of the latest required
    evidence.
    \item Prefer the moment at which the question is most naturally motivated
    by the current scene and activity objective.
    \item When multiple candidates are similarly natural, choose the earliest
    one after all required evidence is available.
    \item The trigger scenario must explain both what is currently happening
    and why a viewer would need to recall the earlier evidence now.
    \item Do not insert the evidence timestamps into the natural-language
    question itself.
\end{enumerate}

\smallskip
\noindent\textbf{Output (JSON only)}

\texttt{\{"t\_query":"\{MM:SS timestamp\}",}\\
\texttt{\quad "trigger\_scenario":"While \{current event\},}\\
\texttt{\qquad a viewer naturally wonders \{information need\}.",}\\
\texttt{\quad "trigger\_event":"\{chosen event description\}"\}}
\end{minipage}
\end{tcolorbox}
\caption{
    Two-stage trigger-assignment prompts. The VLM first retrieves causally valid
    candidate events occurring after all answer evidence is available; the LLM
    then chooses the earliest candidate that most naturally motivates the
    information need and assigns \(t_{\mathrm{query}}\).
}
\label{fig:prompt_trigger}
\end{figure*}

\paragraph{Evaluation Prompts.}
We use separate judge prompts to assess final-answer correctness and reasoning quality. The answer judge compares open-ended predictions against reference answers without video access, whereas the process judge evaluates explicit reasoning traces against temporally ordered gold evidence and the corresponding gold reasoning chain.


\begin{figure*}[t]
\centering
\begin{tcolorbox}[
    colback=dcPromptCream,
    colframe=dcPromptPeach,
    colbacktitle=dcPromptPeach,
    coltitle=black,
    title={\textbf{Prompts for Open-Ended Answer Evaluation}},
    fonttitle=\small,
    fontupper=\scriptsize,
    width=\textwidth,
    boxrule=0.8pt,
    arc=3pt,
]
\raggedright
\begin{minipage}[t]{0.485\textwidth}
\textbf{A.\ Semantic Answer Scoring}

\smallskip
You are evaluating how well a predicted answer matches the ground-truth
answer for a video-understanding question.

\smallskip
\noindent\textbf{Input}

\textbf{Question:} \texttt{\{question\}}\\
\textbf{Ground-truth answer:} \texttt{\{ground\_truth\}}\\
\textbf{Predicted answer:} \texttt{\{prediction\}}

\smallskip
\noindent\textbf{Scoring Guidelines}

\begin{itemize}
    \item For an \textbf{order or sequence}, preserve the ordering; equivalent
    surface forms receive the same score.
    \item For a \textbf{count or number}, require the correct value and assign
    partial credit to near misses.
    \item For a \textbf{duration or time}, score by numerical proximity; a
    difference within approximately \(10\%\) is nearly correct.
    \item For \textbf{objects or descriptions}, assess semantic equivalence
    rather than exact wording.
    \item For a \textbf{comparison}, require the correct direction.
    \item For a \textbf{state, location, or process}, assess whether the key
    facts and details are correctly captured.
\end{itemize}

\smallskip
\noindent\textbf{Score Definition}

\textbf{5}: fully correct; all key facts match semantically.\\
\textbf{4}: nearly correct; the core answer is correct with only minor
omissions or extra detail.\\
\textbf{3}: partially correct; the main idea is relevant but some important
details are missing or wrong.\\
\textbf{2}: loosely related; substantial errors or omissions remain.\\
\textbf{1}: marginally related but largely incorrect.\\
\textbf{0}: contradictory, unrelated, or completely incorrect.

\smallskip
\noindent\textbf{Output}

Return only one integer from 0 to 5, with no explanation.
\end{minipage}
\hfill
\begin{minipage}[t]{0.485\textwidth}
\textbf{B.\ Duration Normalization}

\smallskip
Extract the duration or time value from the following text and convert it to
seconds. If the text contains multiple durations, extract the final answer,
i.e., the main duration being reported.

\smallskip
\noindent\textbf{Examples}

\texttt{17 seconds} \(
ightarrow 17\)\\
\texttt{approximately 2 minutes and 30 seconds} \(
ightarrow 150\)\\
\texttt{about 1.5 minutes} \(
ightarrow 90\)\\
\texttt{around 45s} \(
ightarrow 45\)\\
\texttt{3 minutes} \(
ightarrow 180\)\\
\texttt{1 hour 5 minutes} \(
ightarrow 3900\)

\smallskip
\noindent\textbf{Input}

\textbf{Text:} \texttt{\{text\}}

\smallskip
\noindent\textbf{Output}

Return only a number in seconds, without units or explanation.

\medskip
\hrule
\medskip
\noindent\textbf{Scoring Rule}

Duration-type questions are evaluated separately from semantic answer
scoring. Given a reference duration \(t^*\) and prediction \(\hat{t}\), we
compute
\(
d_{\mathrm{rel}}=|\hat{t}-t^*|/|t^*|
\)
and mark the prediction correct when
\(
d_{\mathrm{rel}}\leq 0.15
\).
\end{minipage}
\end{tcolorbox}
\caption{
    Prompts for open-ended answer evaluation. Semantic answers receive a
    type-aware score from 0 to 5, with scores of at least 4 treated as correct.
    Duration answers are normalized to seconds and evaluated using a 15\%
    relative-error tolerance.
}
\label{fig:prompt_oe_evaluation}
\end{figure*}

\begin{figure*}[t]
\centering
\begin{tcolorbox}[
    colback=dcPromptLilac,
    colframe=dcPromptLavender,
    colbacktitle=dcPromptLavender,
    coltitle=black,
    title={\textbf{Prompt for Evidence Utilization Cascade Evaluation}},
    fonttitle=\small,
    fontupper=\scriptsize,
    width=\textwidth,
    boxrule=0.8pt,
    arc=3pt,
]
\raggedright
\begin{minipage}[t]{0.485\textwidth}
You are evaluating the reasoning process of a video language model for a
multi-hop video QA task. The model watched a long-form video and answered a
question requiring observations from multiple time points.

\smallskip
\noindent\textbf{Task Input}

\textbf{Question:} \texttt{\{question\}}\\
\textbf{Question type:} \texttt{\{qa\_type\}}\\
\textbf{Format:} \texttt{\{fmt\}} (\texttt{\{fmt\_desc\}})\\
\textbf{Model answer:} \texttt{\{pred\_answer\}}\\
\textbf{Gold answer:} \texttt{\{gold\_answer\}}\\
\textbf{Answer correctness:} \texttt{\{correct\_str\}}\\
\textbf{Gold evidence hops:} \texttt{\{gold\_evidence\_text\}}\\
\textbf{Gold reasoning chain:} \texttt{\{gold\_reasoning\}}\\
\textbf{Model think output:} \texttt{\{think\_text\}}

\smallskip
Each gold evidence item follows the format:\\
\texttt{Hop \{i\} [t=\{time\}]: \{object\} -- \{event/action\}}.

\medskip
\noindent\textbf{Per-Hop Analysis}

For every gold evidence hop, evaluate the following cascading properties:

\begin{enumerate}
    \item \textbf{located} (Boolean): Does the think text mention or describe
    the corresponding event or observation anywhere? Count a semantic match;
    exact wording and timestamps are not required.
    \item \textbf{cited} (Boolean or null): Is the located observation actually
    used to derive the answer rather than merely listed? If
    \texttt{located=false}, return \texttt{null}.
    \item \textbf{role\_correct} (Boolean or null): Is the cited observation
    assigned the same logical role as in the gold reasoning chain? Check the
    object or actor, temporal or causal position, counting role, and comparison
    direction. If the hop is not cited, return \texttt{null}.
\end{enumerate}
\end{minipage}
\hfill
\begin{minipage}[t]{0.485\textwidth}
\textbf{Global Reasoning Analysis}

\smallskip
\textbf{Chain completeness} (float from 0 to 1): Estimate the fraction of the
gold reasoning logic reproduced by the think output. Consider whether the key
steps are present, temporal or causal order is preserved, and inter-hop
connections are maintained. A score of 0 denotes no overlap and 1 denotes full
reproduction.

\smallskip
\textbf{Answer faithfulness} (0 or 1): Using only the think output, determine
whether the model answer can be logically derived from the stated observations
and reasoning. Return 0 if the answer contradicts the reasoning, lacks a
sufficient basis, or appears without justification.

\smallskip
\textbf{Visual dependency} (0 or 1): Determine whether the specific answer
requires video observation. Return 0 if it can be obtained from common sense,
question wording, answer options, or world knowledge alone.

\smallskip
\textbf{Hallucination count}: Count the distinct factual claims about the video
in the think output as \texttt{n\_claims}. Count claims unsupported by the gold
evidence, including fabricated events, incorrect objects or states, and
invented temporal relations, as \texttt{n\_hallucinated\_claims}. Exclude vague
or generic statements.

\medskip
\noindent\textbf{Output Format (JSON only)}

\noindent\texttt{\{}\\
\texttt{\quad "per\_hop": [}\\
\texttt{\qquad \{"located": \{bool\}, "cited": \{bool|null\},}\\
\texttt{\qquad\quad "role\_correct": \{bool|null\}\}, ...],}\\
\texttt{\quad "chain\_completeness": \{float in [0,1]\},}\\
\texttt{\quad "answer\_faithfulness": \{0|1\},}\\
\texttt{\quad "visual\_dependency": \{0|1\},}\\
\texttt{\quad "n\_claims": \{integer\},}\\
\texttt{\quad "n\_hallucinated\_claims": \{integer\}}\\
\texttt{\}}

\smallskip
The \texttt{per\_hop} array must contain exactly \texttt{\{n\_hops\}} entries
in gold-hop order. Return no text before or after the JSON object.
\end{minipage}
\end{tcolorbox}
\caption{
    Model-as-judge prompt for process-oriented evaluation of explicit think
    outputs. The judge evaluates evidence localization, citation, and logical
    role per hop, together with chain completeness, answer faithfulness, visual
    dependency, and factual hallucination.
}
\label{fig:prompt_think_evaluation}
\end{figure*}

%% file: files/data_ethics.tex
\section{Data Availability, Licensing, and Ethics}
\label{app:data_ethics}

\paragraph{Open-Source Commitment.}
Upon publication, we will release all artifacts created in this work. The
release will include the complete benchmark annotations, comprising questions,
answers, evidence intervals, query times, task and reasoning labels, and
official data splits, together with the construction prompts, evaluation
prompts, scoring code, and preprocessing scripts. We will also release the
complete implementation of EviStream, including backbone integration,
inference configurations, and reproducibility instructions.

\paragraph{Licensing.}
The benchmark annotations and metadata will be released under the Creative
Commons Attribution-NonCommercial 4.0 International license, while the
EviStream implementation and evaluation code will be released under the Apache
License 2.0. Source videos are drawn from the validation splits of
EPIC-KITCHENS-100, Assembly101, and Ego4D~\cite{damen2022epickitchens,
sener2022assembly101,grauman2022ego4d}. We will not redistribute the original
videos. Users must obtain them from the official providers and comply with the
corresponding licenses and data-use agreements. Backbone weights will likewise
remain subject to their original licenses.

\paragraph{Ethical Considerations.}
This work introduces no new video collection and relies exclusively on
established datasets obtained through their official access procedures. The
benchmark focuses on actions, objects, states, and temporal relations and does
not introduce identity, biometric, or sensitive-attribute annotations.
Model-generated QA pairs undergo human verification for contextual relevance,
answer correctness, evidence sufficiency, and label consistency. Nevertheless,
biases in the source datasets and model-assisted construction process may
remain; we will document data provenance, scenario distributions, annotation
procedures, and known limitations in the release. The benchmark and EviStream
are intended for research on streaming video understanding and should not be
used for surveillance, personal identification, biometric profiling, or
high-stakes decision-making.